%% file: main.tex
\RequirePackage[OT1]{fontenc}
\documentclass[letterpaper, 10 pt, conference]{ieeeconf}
\usepackage{ifthen}
\newcommand\arxivmode{true} 

\ifthenelse{\equal{\arxivmode}{true}}{\pdfoutput=1}{}

\IEEEoverridecommandlockouts
\usepackage{graphicx}
\graphicspath{{./images/}}
\DeclareGraphicsExtensions{.pdf,.png,.jpg}

\let\labelindent\relax
\usepackage{enumitem}
\usepackage{mwe}
\usepackage{amsfonts}
\usepackage{amsmath}
\usepackage{amssymb}
\usepackage{balance}
\usepackage{xcolor}
\usepackage[bookmarks=true]{hyperref}
\usepackage{url}
\usepackage{mathtools}
\usepackage[per-mode=symbol]{siunitx}
\usepackage{subfig}
\usepackage[ruled,linesnumbered]{algorithm2e}
\usepackage[font=small]{caption}
\usepackage{booktabs}
\usepackage{multirow}

\makeatletter
\let\NAT@parse\undefined
\makeatother
\usepackage[numbers,sort&compress]{natbib}
\usepackage[capitalise]{cleveref}

\input{symbols}

\usepackage[utf8]{inputenc}
\usepackage{pgfplots}
\usepgfplotslibrary{groupplots,dateplot}
\usetikzlibrary{patterns,shapes.arrows}
\pgfplotsset{compat=newest}

\usepackage[export]{adjustbox}

\usepackage{tikz}
\usetikzlibrary{external,positioning}
\usepgfplotslibrary{external}
\newcommand{\edit}[1]{{\color{black}{#1}}}
\newcommand{\wennie}[1]{{\color{black}{#1}}}

\crefname{figure}{Fig.}{Figs.}
\Crefname{figure}{Figure}{Figures}
\crefname{table}{Tab.}{Tabs.}
\Crefname{table}{Table}{Tables}
\crefname{algorithm}{Alg.}{Algs.}
\Crefname{algorithm}{Algorithm}{Algorithms}
\crefname{section}{Sect.}{Sects.}
\Crefname{section}{Section}{Sections}

\title{\LARGE \bf
  Hierarchical Collision Avoidance for Adaptive-Speed Multirotor Teleoperation
}
\author{Kshitij Goel, Yves Georgy Daoud, Nathan Michael, and Wennie Tabib
\thanks{The authors are affiliated with the Robotics Institute, Carnegie Mellon University, USA}%
}

\begin{document}
\maketitle


\begin{abstract}
  This paper improves safe motion primitives-based teleoperation of a multirotor
  by developing a hierarchical collision avoidance method that modulates maximum
  speed based on environment complexity and perceptual constraints. Safe speed
  modulation is challenging in environments that exhibit varying clutter.
  Existing methods fix maximum speed and map resolution, which prevents vehicles
  from accessing tight spaces and places the cognitive load for changing speed
  on the operator. We address these gaps by proposing a high-rate
  (\SI{10}{\hertz}) teleoperation approach that modulates the maximum vehicle
  speed through hierarchical collision checking. The hierarchical collision
  checker simultaneously adapts the local map's voxel size and maximum vehicle
  speed to ensure motion planning safety.  The proposed methodology is evaluated
  in simulation and real-world experiments and compared to a non-adaptive motion
  primitives-based teleoperation approach. The results demonstrate the
  advantages of the proposed teleoperation approach both in time taken and the
  ability to complete the task without requiring the user to specify a maximum
  vehicle speed.
\end{abstract}


\section{\label{sec:intro}Introduction}
\input{content/introduction}

\section{\label{sec:related_work}Related Work}
\input{content/related_work}

\section{\label{sec:approach}Technical Approach}
\input{content/approach}

\subsection{\label{ssec:implementation}Implementation Detail}
\input{content/implementation}

\section{\label{sec:results}Results}
\input{content/results}

\section{\label{sec:limitations}Limitations}
\input{content/limitation}

\section{Conclusion}
\input{content/conclusion}

\section{\label{sec:acknowledgments}Acknowledgments}
The authors thank the Mid Atlantic Karst Conservancy for
granting permission to test at a cave on the Barbara Schomer
Cave Preserve. The authors also thank D. Wettergreen
and S. Vats for their feedback on this manuscript.

\balance


{
  \footnotesize
  \bibliographystyle{unsrtnat}
  \bibliography{refs,bibliography/kshitij_library}
}

\end{document}

%% file: symbols.tex
\newcommand{\freespace}{\mathcal{X}_{\mathrm{free}}}
\newcommand{\surfacespace}{\mathcal{X}_{\mathrm{occ}}}
\newcommand{\unknownspace}{\mathcal{X}_{\mathrm{unk}}}
\newcommand{\unsafespace}{\mathcal{X}_{\mathrm{unsafe}}}
\newcommand{\thdspace}{\mathbb{R}^3}

\newcommand{\body}{\mathcal{B}}

\newcommand{\state}{\mathbf{x}}

\newcommand{\controlstate}{\state^{c}}

\newcommand{\joyinput}{\mathbf{a}}

\newcommand{\prim}{\gamma}
\newcommand{\selprim}{\gamma_{\mathrm{sel}}}
\newcommand{\stopprim}{\gamma_{\mathrm{stop}}}
\newcommand{\selvel}{V_{x}}
\newcommand{\zmaxvel}{V_{z}}
\newcommand{\omaxvel}{\Omega}
\newcommand{\primdur}{T}
\newcommand{\accmax}{A_{x}}

\newcommand{\plantime}{\Delta t_{p}}
\newcommand{\maptime}{\Delta t_{m}}
\newcommand{\sensetime}{\Delta t_{s}}
\newcommand{\totallatency}{\Delta t_{l}}

\newcommand{\mapres}{\alpha}
\newcommand{\delres}{\Delta \alpha}
\newcommand{\prevres}{\alpha_{\mathrm{prev}}}
\newcommand{\maxres}{\alpha_{\max}}
\newcommand{\minres}{\alpha_{\min}}

\newcommand{\changelevel}{\Delta l}
\newcommand{\map}{\mathbf{m}}
\newcommand{\distancemap}{\mathbf{d}}
\newcommand{\totalvoxels}{|\map|}
\newcommand{\xvoxels}{N_x}
\newcommand{\yvoxels}{N_y}
\newcommand{\zvoxels}{N_z}
\newcommand{\bbx}{\mathbf{B}}
\newcommand{\bbxcorner}{\mathbf{b}}
\newcommand{\keyframe}{\mathcal{K}}
\newcommand{\keyframethres}{\beta}
\newcommand{\equivdist}{z_{\mathrm{eq}}}
\newcommand{\delv}{\delta v}

\newcommand{\meas}{\mathcal{Z}}

\newcommand{\sensorrange}{z_{\max}}
\newcommand{\collisionradius}{r_{\mathrm{coll}}}
\newcommand{\robotradius}{r_{\mathrm{robot}}}


%% file: content/introduction.tex
\begin{figure}[!t]
  \centering
  \ifthenelse{\equal{\arxivmode}{true}}
  {\subfloat[\label{sfig:cave_entrance}]{\includegraphics[width=0.49\linewidth,trim=0 0 172 88,clip]{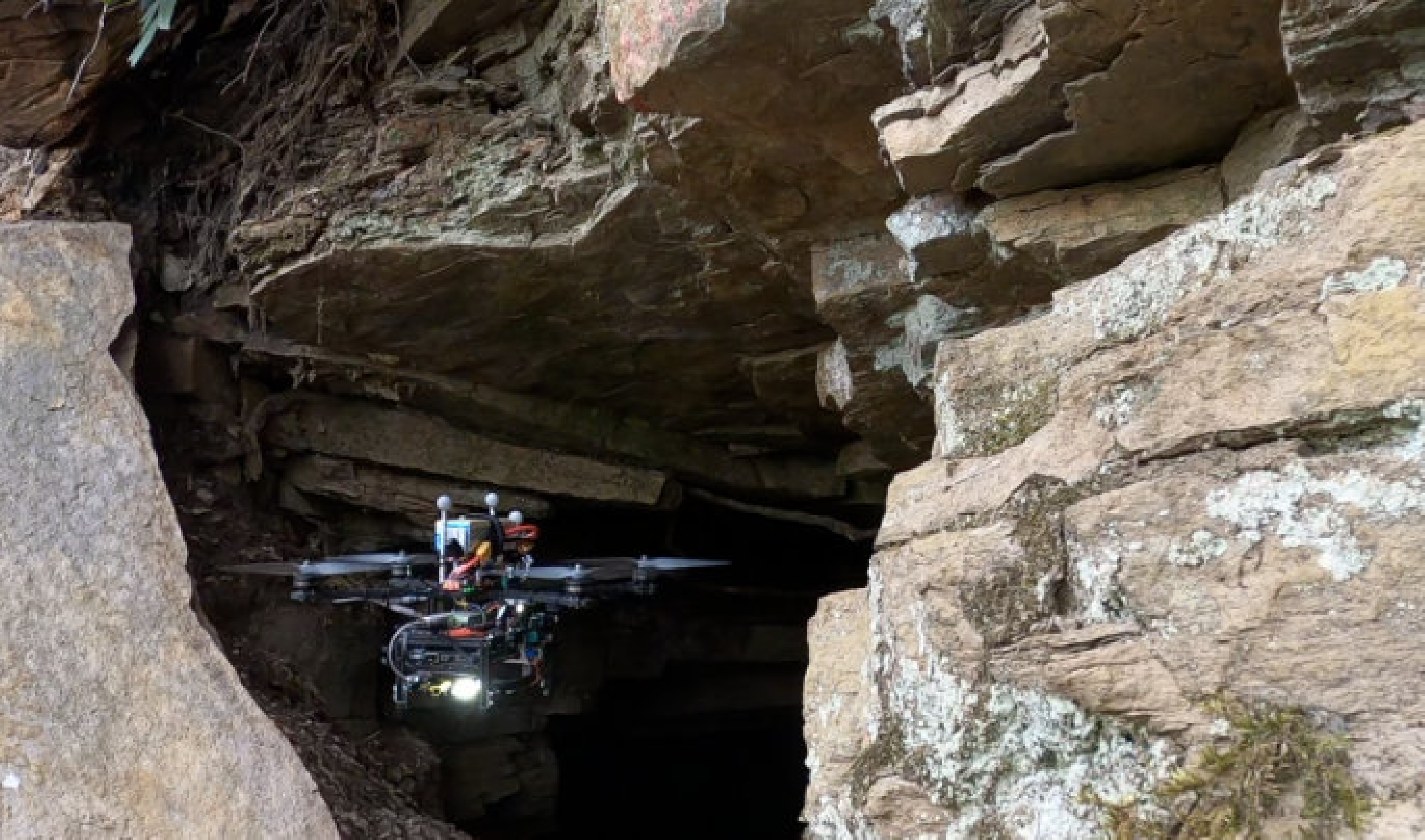}}}
  {\subfloat[\label{sfig:cave_entrance}]{\includegraphics[width=0.49\linewidth,trim=0 0 172 88,clip]{images/glory_shot/robot_enters_cave_compressed2.eps}}}%
  \ifthenelse{\equal{\arxivmode}{true}}
  {\subfloat[\label{sfig:cave_passage}]{\includegraphics[width=0.49\linewidth,trim=150 30 150 120,clip]{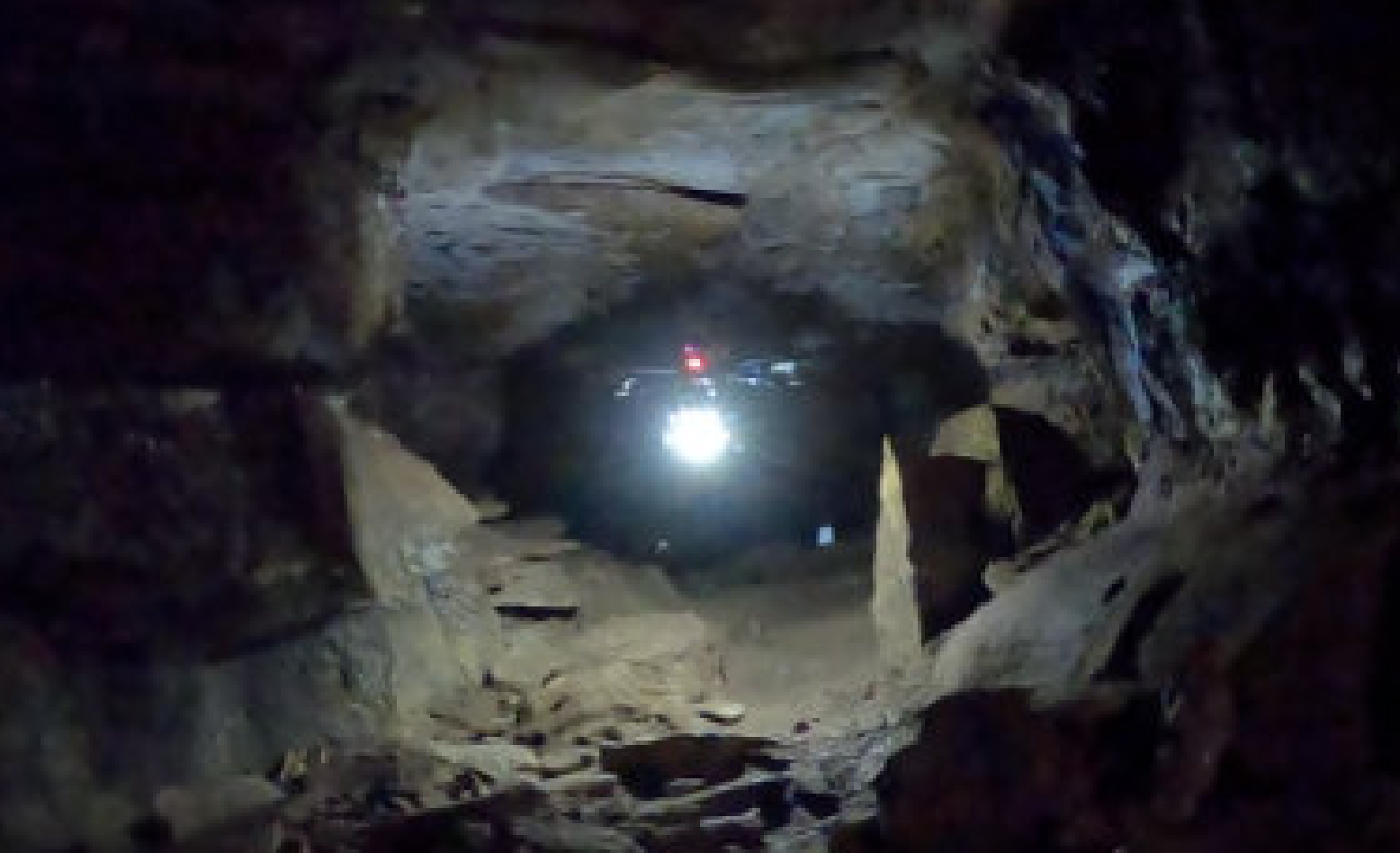}}}
  {\subfloat[\label{sfig:cave_passage}]{\includegraphics[width=0.49\linewidth,trim=150 30 150 120,clip]{images/glory_shot/robot_in_tunnel-compressed.eps}}}\\
  \ifthenelse{\equal{\arxivmode}{true}}
  {\subfloat[\label{sfig:cave_entrance2}]{\includegraphics[width=0.98\linewidth,trim=0 200 0 200,clip]{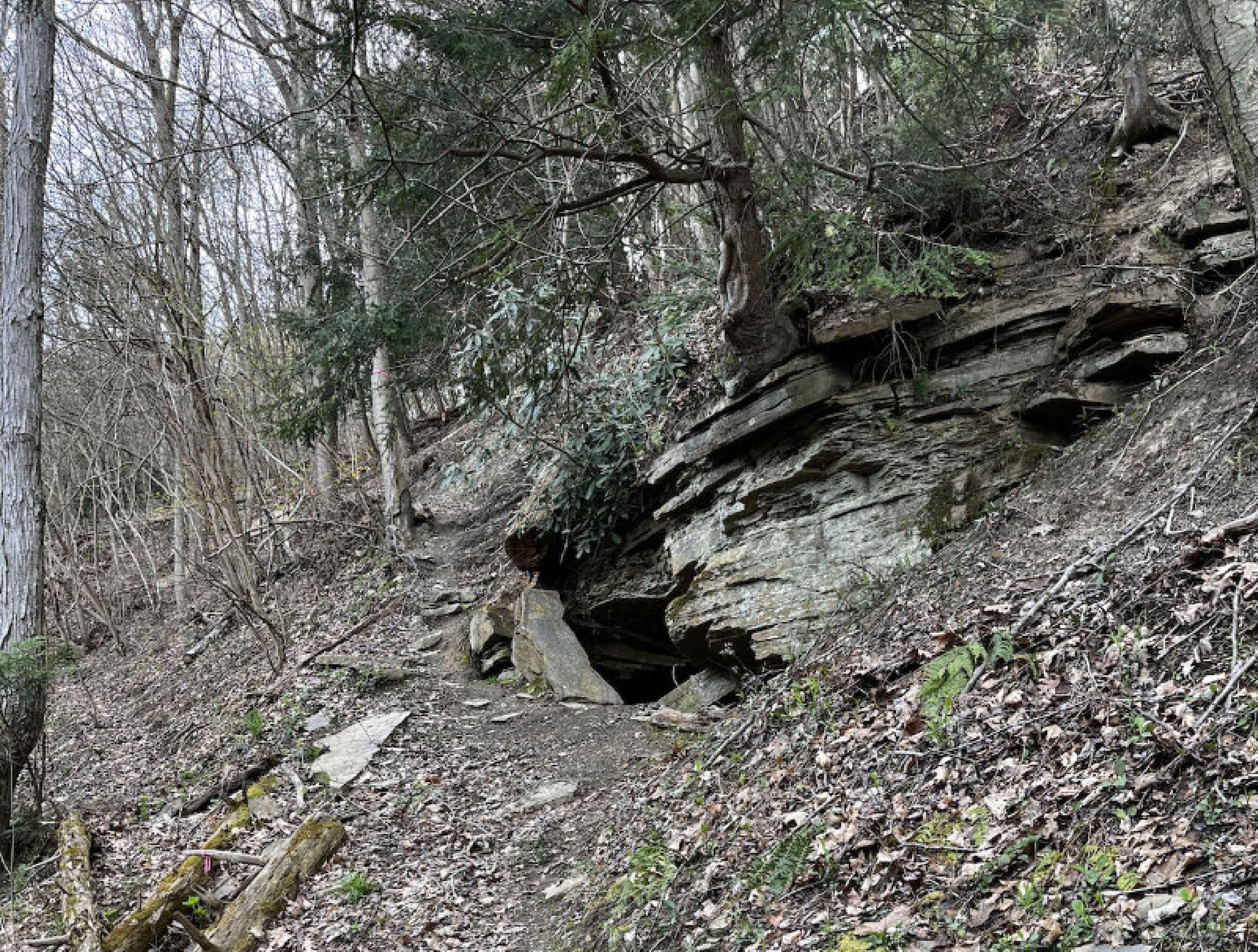}}}
  {\subfloat[\label{sfig:cave_entrance2}]{\includegraphics[width=0.98\linewidth,trim=0 200 0 200,clip]{images/glory_shot/iphone_sf_entrance-compressed.eps}}}
  \caption{\label{fig:challenges}
    Tele-operated multirotor adapts the motion planning speed and local map resolution
    to~\protect\subref{sfig:cave_entrance} enter a cave and~\protect\subref{sfig:cave_passage}
    traverse a tight passage inside.~\protect\subref{sfig:cave_entrance2} illustrates the
    surroundings near the cave entrance, which is embedded in a sloping
    hillside. A video of this experiment can be found at \url{https://youtu.be/VjyoPVXT8WY}.}
\end{figure}

During cave search and rescue (CSAR), the Initial
Response Team (IRT) executes a ``hasty'' search of the cave to provide
rapid situational awareness (e.g., determining the probability of a victim's
location in various passages in the cave)~\citep{hempel_call_2001}.
This team may not have access to a prior map of
the cave (e.g., there are 4378 documented caves in Virginia and only 1348 have
been mapped) which makes searching for victims underground
challenging~\citep{stella-watts_epidemiology_2012}. Inaccessible areas (e.g.,
vertical ascents or tight passages) further complicate the situational awareness
task. We envision teleoperated multirotor vehicles integrated into the IRT to
rapidly search for victims, enhance the accessibility of the cave, and reduce
the time taken to gain situational awareness.  For example, a member of the IRT
should be able to rapidly and safely teleoperate a multirotor through narrow passages.
A key challenge for rapid CSAR is to reduce the cognitive
load on the operator. To address this challenge, we develop a teleoperation methodology that
adjusts the robot speed automatically based on environment complexity and perceptual constraints.
Speed modulation in cave domains can be challenging due to varying
clutter -- open spaces can allow for high speeds but narrow passages require slow
speeds to maintain safety. Allowing safety through narrow passages requires maintaining
a local environment map and onboard collision checking
assistance~\citep{spitzer_fast_2020}. Depending on the resolution of the local
environment map, narrow passages may or may not be visible to the collision
checker. Thus, for teleoperation through narrow passages, variable-resolution
local environment maps are required. Towards meeting the speed modulation and
variable-resolution mapping requirements, the contributions of this paper are:
(1) A hierarchical collision avoidance method that enables maximum speed
modulation in a motion primitives-based teleoperation framework through updates
to the map resolution and (2) evaluation of the method through simulated and real
world experiments using a quadrotor, including one experiment in a wild cave
(\cref{fig:challenges}).

%% file: content/related_work.tex
Motion primitives-based teleoperation of a multirotor has been demonstrated to
reduce operator cognitive load~\citep{yang_framework_2017}.
~\citet{spitzer_fast_2020} propose a teleoperation approach that
leverages a KD-Tree to locally represent the
environment for fast collision checking.  A motion primitive pruning approach
leverages low-latency collision avoidance and the closest distance
to the operator's joystick input to allow for adaptive-speed teleoperation in
the presence of obstacles. The local environment representation assumes the
resolution of the map is fixed before the vehicle is teleoperated
and does not encode unknown or free space information. However,
in practice, the unknown space information may be necessary to ensure safety when
the environment is incrementally revealed through a limited range depth
sensor~\citep{janson_safe_2018,tordesillas_faster_2021}. Free space information
is necessary to simultaneously support applications such as robotic
exploration~\citep{tabib_autonomous_2021}.  Thus, in this work, we use
probabilistic occupancy maps for local environment representation.
A
fixed resolution during operation imposes restrictions on the configuration
space for the planner because the motion primitive pruning approach depends on
the granularity of discretization. If the resolution is too low, the robot might
not be able to enter narrow entrances due to the coarseness of the map. If the
resolution is too high, the high perceptual latency of a fine map limits the
maximum speed of the robot~\citep{falanga_how_2019}. In this work, we address
this research gap by proposing a multirotor teleoperation approach, which uses
variable-resolution local probabilistic occupancy maps hierarchically
to enable fast teleoperation in open spaces while ensuring safe, low-speed
teleoperation through narrow passages.

Prior work in adaptive motion planning has been proposed to
modulate robot speed based on application-specific
heuristics.~\citet{zhang_falco_2020} present a likelihood-based collision
avoidance strategy for fast teleoperation of a multirotor by
prioritizing open spaces for navigation to maintain high
speed. The objective assumes that taking an alternate path (i.e., a
path through open space) will lead to the same location as taking another (i.e.,
a path through a narrow passage); however, this assumption is flawed in the
context of certain domains like caves. In contrast, our work adapts
the maximum vehicle speed according to environment complexity.
\citet{quan_eva-planner_2021} propose an adaptive optimization-based motion
planning approach for the multirotor navigation task, which is most
similar to our approach. The heuristic used for speed modulation relies on the
angle between the velocity direction and the gradient of the local signed
distance field. This heuristic is coupled with a multi-layer model predictive
control approach to allow for fast flight in sparsely cluttered environments and
slow flight through dense clutter.  However, while the motion planner allows
adaptation in speed, the resolution of the environment representation is fixed.
Selecting the resolution of an \emph{a priori} unknown
environment is difficult, which makes the approach
of~\citet{quan_eva-planner_2021} unsuitable for our application. An alternative
strategy is to operate with a range of voxel sizes for the local
environment representation and modulate speeds according to the change in the
map. The proposed approach builds on this strategy and introduces a motion
primitive selection approach that modulates the maximum speed along the motion
primitives according to the voxel size of the local occupancy map.

Several methods exist for hierarchical volumetric occupancy mapping.  OctoMap
by~\citet{hornung_octomap_2013} provides a multi-level representation of
occupancy via an OcTree data structure. However, to the best of our knowledge,
no motion planner exists that can leverage multiple levels of a local
OctoMap-based representation to allow for adaptive-speed teleoperation through
environments with varying clutter.~\citet{nelson_environment_2018} present an
occupancy grid adaptation
methodology for two-dimensional environment exploration using ground robots.
This approach is specific to the exploration scenario and the voxel sizes are
adapted via the information-bottleneck method to minimize the cost of computing
the exploration objective. It is unclear how this approach may be applied in the
context of multirotor teleoperation since the objective highly depends on the
intent of the human operator and may vary over time. Closest to our
work,~\citet{funk_multi-resolution_2021} create a multi-resolution OcTree-based
representation for accurate 3D reconstruction and online
planning. The resolution is selected by minimizing the error in occupancy
representation. A path planning result is presented in this work that utilizes a
``coarse-to-fine" approach for collision checking. The time to compute the
motion plan varies from $\SI{0.01}{\second}$ to $\SI{0.4}{\second}$ depending on
the desired start and end-points and the maximum resolution set by the user.
However, the results are generated with post-processed datasets and a
parallelized implementation so it is unclear how the performance would translate
to compute constrained robotic systems and real-world results.  A key
consideration in the teleoperation context is to minimize the lag in the
multirotor response felt by the operator by having a high motion planning rate.
Thus, it is important to design a planning approach that maintains a steady
planning rate in environments with varying clutter. The proposed approach
addresses this gap by adapting a pre-specified number of map levels per
planning round to achieve a planning rate of at least~\SI{10}{\hertz}.

%% file: content/approach.tex
\begin{figure}
  \centering
  \includegraphics[width=\linewidth]{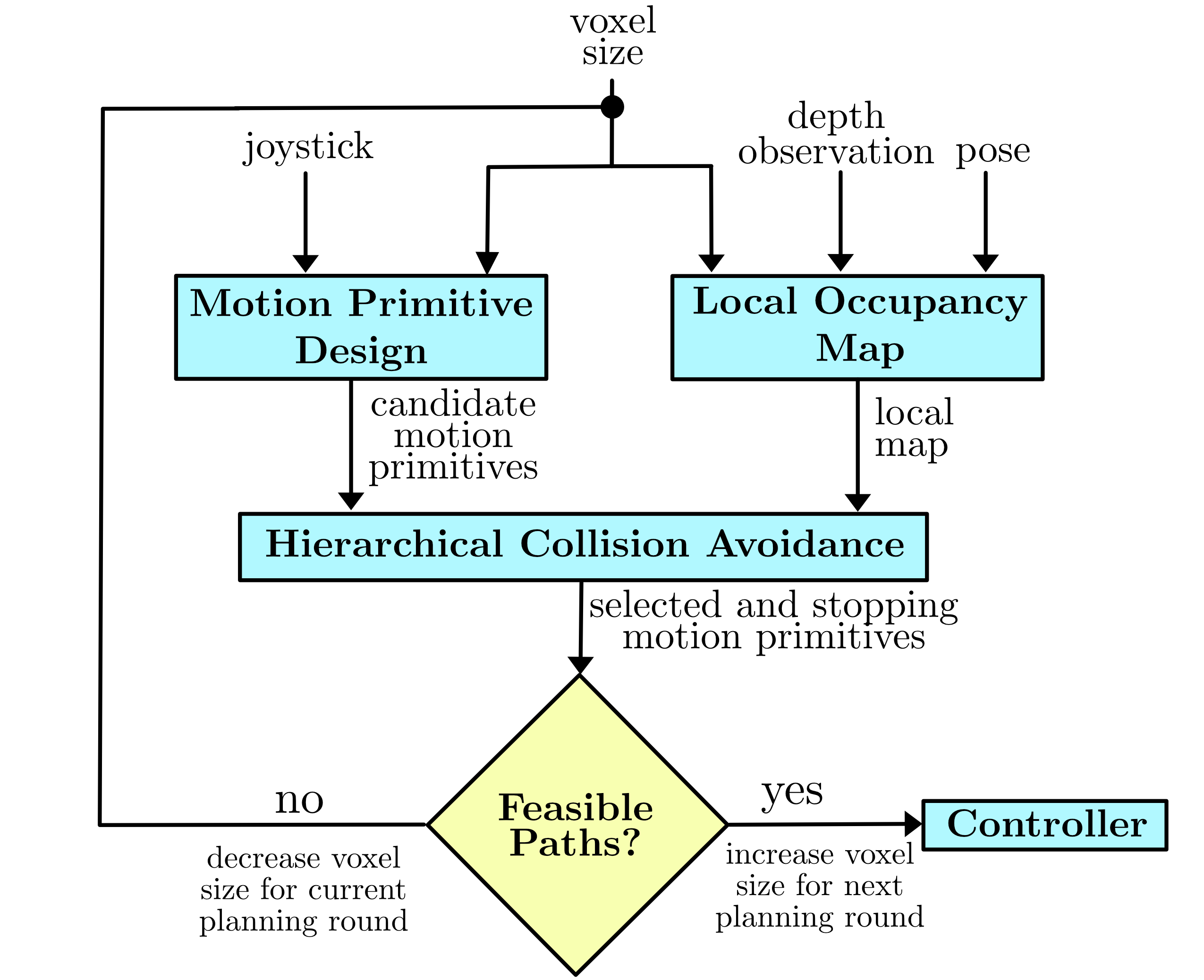}
  \caption{\label{fig:flowchart} Information flow diagram for the technical approach.}
\end{figure}

\begin{figure*}
  \centering
  \subfloat[\label{sfig:bbox1}]{\includegraphics[width=0.33\linewidth,trim=500 100 400 30,clip]{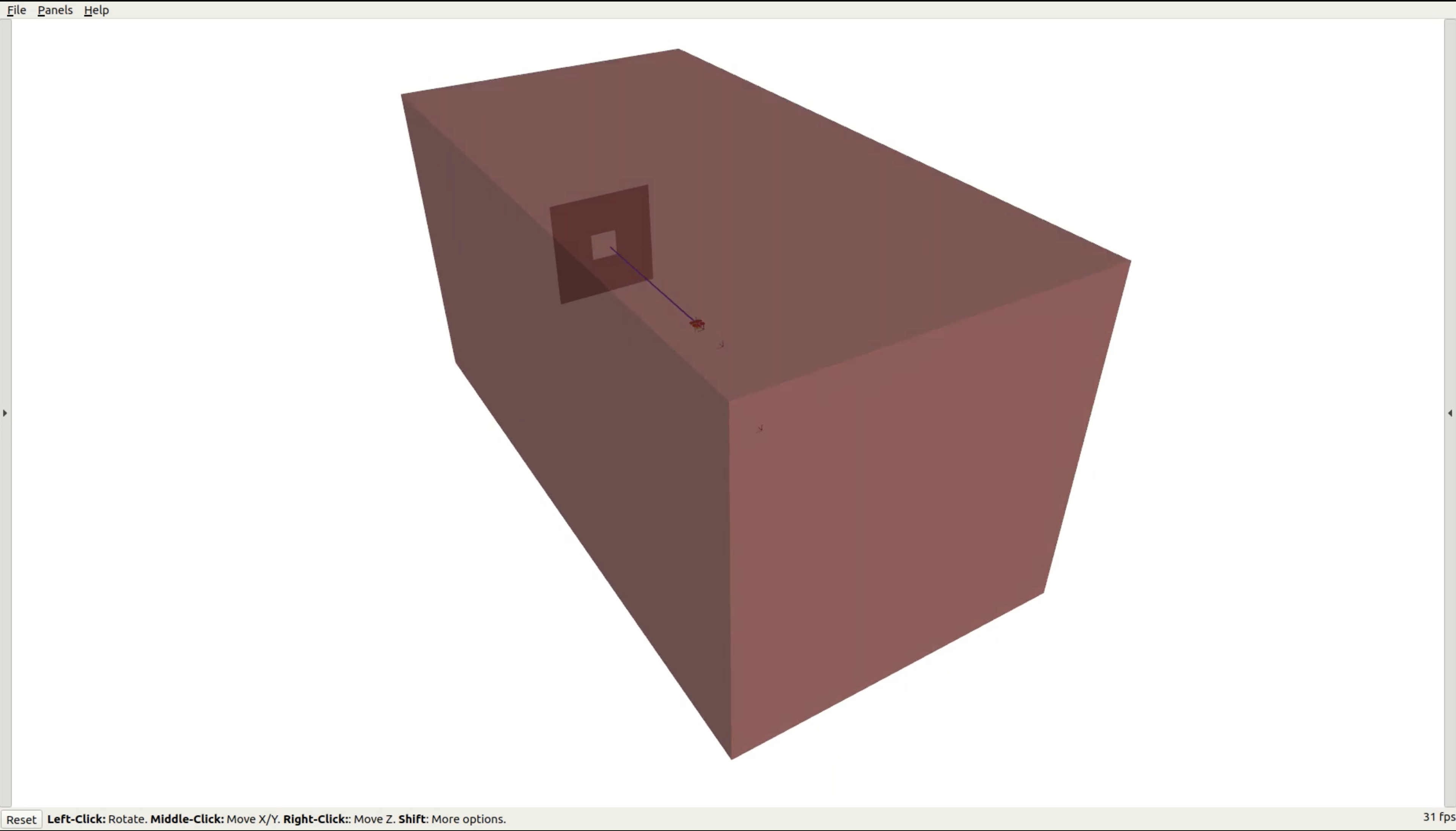}}%
  \subfloat[\label{sfig:bbox2}]{\includegraphics[width=0.33\linewidth,trim=500 100 400 30,clip]{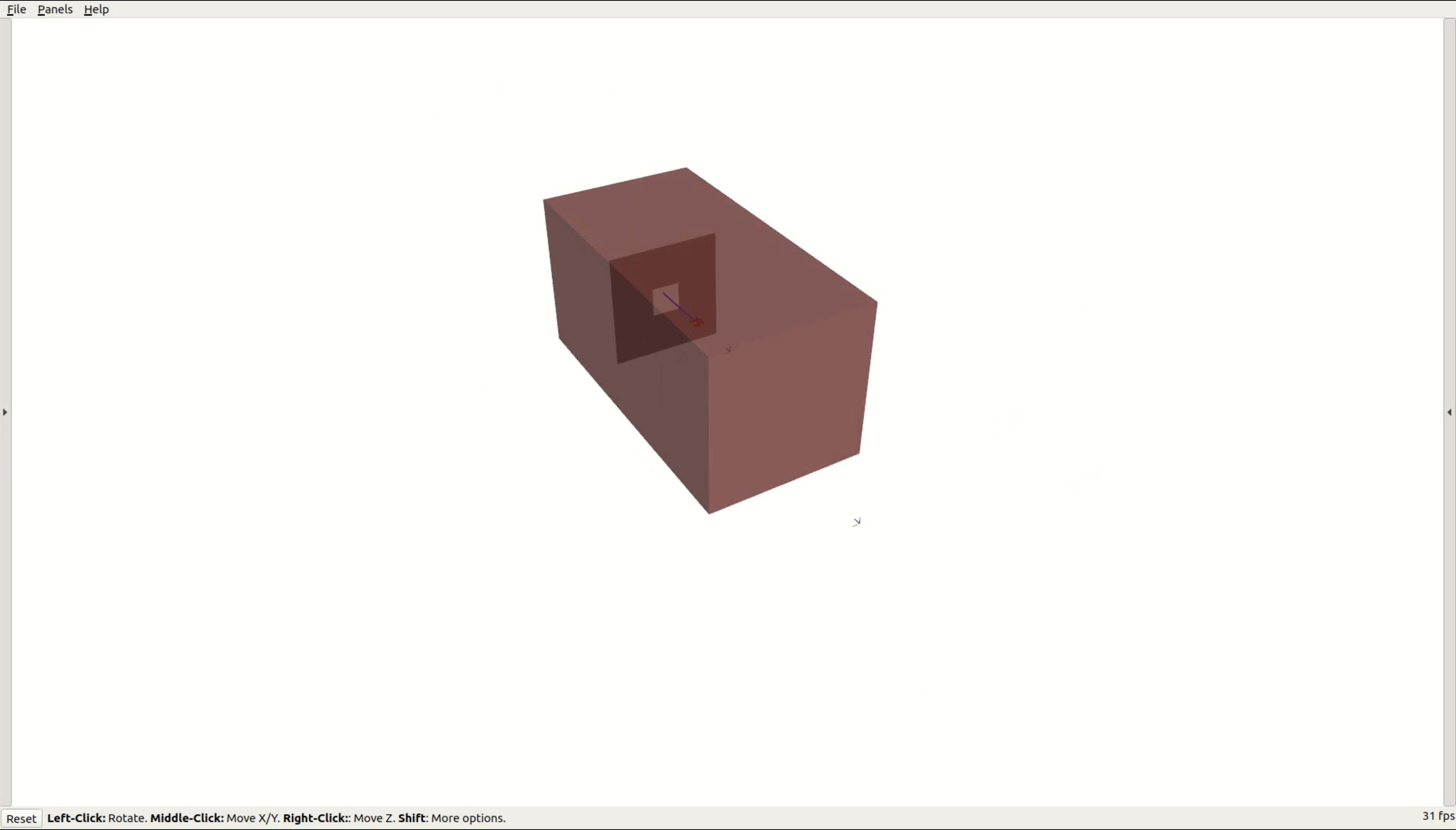}}%
  \subfloat[\label{sfig:bbox3}]{\includegraphics[width=0.33\linewidth,trim=500 100 400 30,clip]{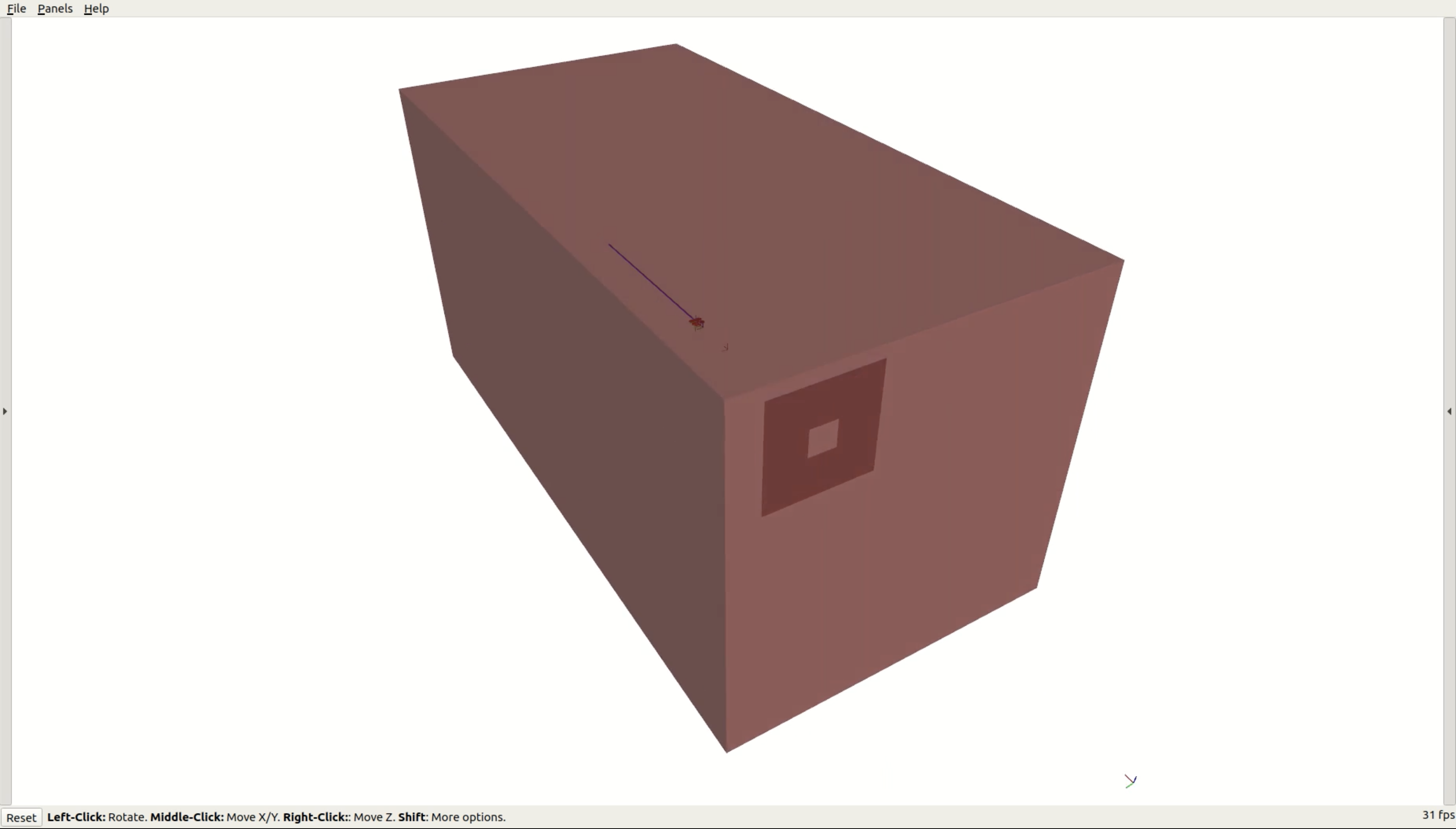}}
  \caption{\label{fig:bbx-adapt}
      Bounding box extents for a scenario where the robot traverses a window.
      The teleoperator gives maximum joystick input in the forward direction for
      these three figures.~\protect\subref{sfig:bbox1} When the robot is far
      from the window, the bounding box extents and local occupancy map are
      large because the voxel size is also large.~\protect\subref{sfig:bbox2}
      As the multirotor gets closer to the window, the voxel size decreases and
      so does the bounding box extent because the number of voxels in the map
      stays the same.~\protect\subref{sfig:bbox3} After exiting the window, the
      bounding box expands to the original size. Note that the change in
      bounding box extents is achieved by varying the voxel size and keeping
      the number of voxels constant.}
\end{figure*}

This section details the adaptive teleoperation approach.~\Cref{fig:flowchart}
illustrates the components of the approach and how information flows between
components.  A local occupancy map is generated using depth observations, robot
pose and voxel size (\cref{ssec:local_mapping}). Note that due to the myopic
nature of the teleoperation task, only a few recent observations are required to
be included in the local occupancy map. The motion primitive design block takes
input from the operator's joystick and the voxel size of the local occupancy
map. This information is used to compute a candidate motion primitive close to
the operator's input and a fallback stopping motion primitive to use in case the
next planning round fails~(\cref{ssec:mp_generation}). The hierarchical
collision avoidance block takes as input these two motion primitives and the
local map to perform collision checking~(\cref{ssec:hier_coll_avd}). If both
selected and stopping motion primitives are found to be feasible, they are sent
to the controller for execution and the voxel size is increased for the next
planning round. In case either of these primitives is infeasible, then the
voxel size is decreased (and therefore the occupancy grid extents,
see~\cref{fig:bbx-adapt}) and the process is repeated for the same planning
round. If a suitable plan is not found, the stopping action from the previous
planning round is executed.

\subsection{\label{ssec:local_mapping}Variable Resolution Local Occupancy Mapping}
A three-dimensional local occupancy grid map $\map$
is generated in the body frame $\body = \{ x_{\body}, y_{\body}, z_{\body} \}$
of the robot using the latest control state $\controlstate$ as the origin, voxel
size $\mapres$, and number of voxels $\totalvoxels = \xvoxels \times
\yvoxels \times \zvoxels$.  The number of voxels along each dimension, $\{
\xvoxels, \yvoxels, \zvoxels \}$, are fixed throughout teleoperation but the
voxel size $\mapres$ may vary during or across planning rounds. The local
occupancy grid bounding box $\bbx = \{ \bbxcorner_{\min}, \bbxcorner_{\max} \}$
extents are adjusted as a function of the voxel size, $\mapres$.
$\bbxcorner_{\min} = \{ x_{\min}, y_{\min}, z_{\min} \}$ is the minimum
$x_{\body}-y_{\body}-z_{\body}$ coordinate of the bounding box and
$\bbxcorner_{\max} = \{ x_{\max}, y_{\max}, z_{\max} \}$ is the maximum
$x_{\body}-y_{\body}-z_{\body}$ coordinate of $\bbx$. Each voxel $m_i \in \map$
is a Bernoulli random variable whose value is $0$ if it is free and $1$ if it is
occupied. Initially, the map is set to have a uniform occupancy probability in
all voxels, $p(m_i) = 0.5, \forall \, i \in \{1, \ldots, \totalvoxels\}$.

It is assumed that the multirotor is equipped with a limited field-of-view (FoV)
forward-facing depth camera that provides a dense depth measurement $\meas$ at a
user-specified sensing rate $1 / \sensetime$. In addition to the latest
state-measurement pair (``keyframe''), $\keyframe_l = \{ \controlstate_l,
\meas_l \}$, one past keyframe is maintained $\keyframe_p = \{ \controlstate_p,
\meas_p \}$. The past keyframe is selected based on a Euclidean distance
threshold $\keyframethres$ between the latest state $\controlstate_l$ and the
previous keyframe state $\controlstate_p$. Measurement $\meas_p$ is transformed
into the frame of reference of the keyframe $\keyframe_l$. The occupancy map is
updated using both $\meas_l$ and the transformed $\meas_p$ through the standard
logodds update~\citep{thrun_probabilistic_2002}.  If the voxel size $\mapres$
changes during the planning round, the occupancy grid map is regenerated using
$\meas_l$ and $\meas_p$ using the new $\mapres$. The time complexity of this
regeneration step depends on the total number of voxels. In practice, we decide
the number of voxels based on the available compute and keep them fixed
throughout teleoperation.

The local occupancy map $\map$ partitions the space
$\thdspace$ into three subspaces: (1) free space $\freespace$, (2) occupied space
$\surfacespace$, and (3) unknown space $\unknownspace$.  To ensure safety, the
motion plans sent to the robot must lie in free space, $\state_t \in
\freespace$, for all time $t$. To check the motion plans against $\unsafespace =
\{ \surfacespace \, \cup \, \unknownspace \}$, a common strategy is to compute a
discrete distance field $\distancemap$ where each point in the field stores the
shortest distance to $\unsafespace$. We use the variant of the fast-marching method
by~\citet{sethian_fast_1996} over occupancy grids to compute the distance map
$\distancemap$ from the local map $\map$. It is assumed that the robot can fit a cube of
side-length $2 \cdot \robotradius$ and a $\collisionradius$ amount of tolerance from
$\unsafespace$ is required for safety.

\subsection{\label{ssec:mp_generation}Motion Primitive Design}
We use the control input parameterization
by~\citet{yang_framework_2017} that maps the space of joystick inputs
to a finite set of forward-arc motion primitives. The joystick input is represented as
$\joyinput_t = \{ v_{x,t}, v_{z,t}, \omega_{t} \}$, where $v_{x,t}$ is the
velocity command in the $x_{\body}$ direction, $v_{z,t}$ is the velocity command
in the $z_{\body}$ direction, and $\omega_t$ is the angular velocity command
around the $z_{\body}$ direction.  All velocity commands are uniformly dense
sets clamped with user-specified bounds: $v_{x,t} \in [-\selvel, \selvel]$,
$v_{z,t} \in [-\zmaxvel, \zmaxvel]$, and $\omega_t \in [-\omaxvel, \omaxvel]$.
Assuming a user-specified duration of the motion primitive, $\primdur$, a
forward-arc motion primitive $\prim_t = \{ \joyinput_t, T \}$ can be generated
by propagating the unicycle model~\citep{pivtoraiko_autonomous_2009}. The motion
primitive $\prim_t$ is checked for feasibility and sent to the controller for execution.

Prior motion primitives-based teleoperation frameworks that utilize forward-arc
motion primitives assume velocity command bounds to be constant and
user-specified~\citep{yang_framework_2017,spitzer_fast_2020}. These bounds can
influence the design of the motion primitives when they are used for navigation
in unknown environments. Such design decisions are made to ensure
that the robot never enters an inevitable collision
state~\citep{janson_safe_2018}. For example, setting the $x_{\body}$-velocity
bound, $\selvel$, depends on many factors: (1) mapping time ($\maptime$), (2)
planning time ($\plantime$), (3) sensing time ($\sensetime$), (3) sensing range
($\sensorrange$), (4) collision tolerance distance ($\collisionradius$), (5)
robot radius ($\robotradius$), and (6) maximum $x_{\body}$-deceleration
($\accmax$)~\citep{goel_fast_2021}. For the teleoperation task, we also need to
account for the bounding box $\bbx$ of the local occupancy map $\map$. In terms
of the voxel size and the number of voxels, the corners of the local map
bounding box $\bbx$ are given by $\bbxcorner_{\min} = - (\mapres / 2) \cdot \{
\xvoxels, \yvoxels, \zvoxels \}$ and $\bbxcorner_{\max} = (\mapres / 2) \cdot \{
\xvoxels, \yvoxels, \zvoxels \}$. Since the robot is at the center of $\bbx$,
the map information available in front of the robot is up to a distance
$\equivdist = \min(\sensorrange, (\mapres / 2) \xvoxels)$ from the robot.
Thus, an ideal upper bound for the maximum $x_{\body}$-velocity is derived
using Euler motion equations as:
\begin{align}
    \selvel = \accmax \left( \sqrt{ \totallatency^2 + 2
    \frac{\equivdist - (\robotradius +
    \collisionradius)}{\accmax} } - \totallatency \right),
    \label{eq:vel_bound_teleop}
\end{align}
where, $\totallatency = \sensetime + \maptime + 2\plantime$. This equation represents an ideal upper
bound because it assumes the deceleration $\accmax$ is attained instantly and
does not consider the motor dynamics. Therefore, in practice, we reduce
$\selvel$ with a constant $\delv$ to account for these unmodeled factors.

Note that~\cref{eq:vel_bound_teleop} represents the velocity bound in terms
of the voxel size $\mapres$ of the local occupancy map $\map$. Thus, the
velocity bound should scale according to the resolution of the map.
Consequently, the motion primitive design is dependent on the voxel size. We use
this fact to adapt voxel size and the motion primitive design across and within
planning rounds through hierarchical collision avoidance
(\cref{ssec:hier_coll_avd}).

\begin{algorithm}
    \DontPrintSemicolon

    \SetKwInput{Input}{input}
    \SetKwInput{Parameters}{parameters}
    \SetKwInput{Output}{output}

    \SetKwProg{Fn}{function}{}{end}

    \SetKwFunction{HierarchicalCollisionAvoidance}{HCA}
    \SetKwFunction{LocalMap}{LocalMap}
    \SetKwFunction{DistanceMap}{DistanceField}
    \SetKwFunction{ComputeMaxVelocity}{MaxSpeed}
    \SetKwFunction{MapJoystick}{MapJoystick}
    \SetKwFunction{ComputeStoppingAction}{StoppingAction}
    \SetKwFunction{CollisionCheck}{InCollision}
    \SetKwFunction{GetPlannerState}{GetPlannerState}
    \SetKwFunction{SchedulePrimitives}{SchedulePrimitives}
    \Fn{\HierarchicalCollisionAvoidance{$\state_{t + \plantime}$, $\joyinput$, $\prevres$}}{
        \Input{$\state_{t + \plantime}$, $\joyinput$, $\prevres$}
        \Parameters{$\maxres$, $\minres$, $\delres$, $\plantime$, $\maptime$, $\sensetime$, $\xvoxels$, $\yvoxels$, $\zvoxels$}
        \Output{$\selprim$, $\stopprim$, $\prevres$}
        \BlankLine
        $\mapres \leftarrow \min(\max(\prevres + \delres, \minres), \maxres)$\;\nllabel{line:start_res}
        $\changelevel \leftarrow 0$\;
        \While{$\changelevel \leq 2$}{
            $\selvel \leftarrow$ \ComputeMaxVelocity($\map$, $\plantime$, $\maptime$, $\sensetime$)\nllabel{line:max_vel}\;
            $\selprim \leftarrow$ \MapJoystick($\state_{t + \plantime}$, $\joyinput$, $\selvel$)\nllabel{line:map_joy}\;
            $\stopprim \leftarrow$ \ComputeStoppingAction($\selprim$, $\plantime$)\nllabel{line:stopping_action}\;
            $\map \leftarrow$ \LocalMap($\mapres$, $\xvoxels$, $\yvoxels$, $\zvoxels$)\nllabel{line:local_map}\;
            $\distancemap \leftarrow$ \DistanceMap($\map$)\nllabel{line:distance_map}\;
            \eIf{\CollisionCheck($\selprim$, $\stopprim$, $\distancemap$)}{\nllabel{line:coll_check}
                $\mapres \leftarrow \min(\max(\mapres - \delres, \minres), \maxres)$\;\nllabel{line:update_res}
                $\changelevel \leftarrow \changelevel + 1$\;
            }{
                break\;
            }
        }
        $\prevres \leftarrow \mapres$\;\nllabel{line:store_res}
        \eIf{$\changelevel > 2$}{\nllabel{line:level_check}
            \Return{$\prevres$}\nllabel{line:return_res_only}
        }{
            \Return{$\selprim$, $\stopprim$, $\prevres$}\nllabel{line:planning_success}
        }
    }
    \caption{Hierarchical Collision Avoidance}\label{alg:hierarch_coll_avd}
\end{algorithm}

\subsection{\label{ssec:hier_coll_avd}Hierarchical Collision Avoidance}
\Cref{alg:hierarch_coll_avd} shows the pseudocode for the Hierarchical Collision
Avoidance (HCA) algorithm.  Instead of using a fixed voxel size $\mapres$ for
the local occupancy map $\map$ throughout teleoperation, it is adapted hierarchically
based on the output of the collision checker. At the start
of the planning round, the voxel size $\mapres$ is set $\delres$ above the one
used in the previous planning round, $\prevres$, but clamped by a pre-specified
$\maxres$ and $\minres$ (\cref{line:start_res}). The motion primitive design is
updated for this voxel size (\cref{ssec:mp_generation}, \cref{line:max_vel}).
The joystick input is mapped to the closest motion primitive, $\selprim$, via
grid search~\citep{yang_framework_2017} (\cref{line:map_joy}).  A stopping
motion primitive, $\stopprim$, is generated as a fallback action in case the
next planning round fails (\cref{line:stopping_action}).  The distance field
(\cref{ssec:local_mapping}, \cref{line:distance_map}) generated via the local
map at the voxel size $\mapres$ (\cref{ssec:local_mapping},
\cref{line:local_map}) is used for collision checking. If either $\selprim$ or
$\stopprim$ are in collision (\cref{line:coll_check}), we reduce the voxel size
by $\delres$ (\cref{line:update_res}), and try planning again. The map
generation step contributes the most to the time complexity of this algorithm.
If a feasible plan is found within a maximum of three map level changes
(\cref{line:level_check}), it is returned (\cref{line:planning_success}) and the
next plan is executed. We choose to limit the checks to three map levels at a
time to keep a consistent time complexity across planning rounds. If a feasible
plan is still not possible, the current voxel size is returned as $\prevres$
(\cref{line:return_res_only}) to use in the next planning round, and the
previously planned stopping action is executed.

%% file: content/implementation.tex
\begin{figure}[!t]
    \centering
    \ifthenelse{\equal{\arxivmode}{true}}
    {\includegraphics[width=\linewidth,trim=0 130 0 150,clip]{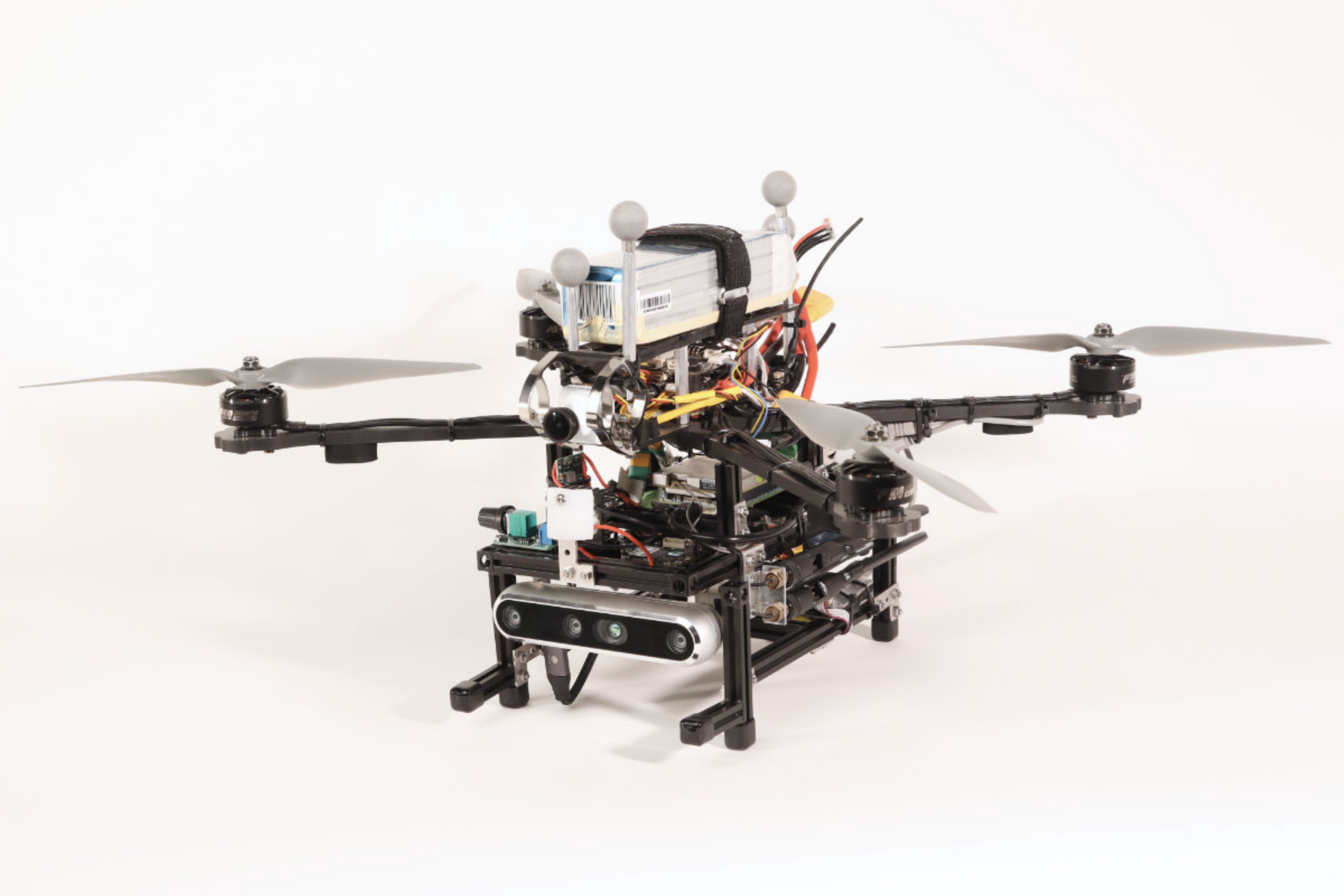}}
    {\includegraphics[width=\linewidth,trim=0 130 0 150,clip]{./images/omicron01-compressed.eps}}
    \caption{\label{fig:omicron01} The robot used in the field experiments is
    equipped with a forward-facing Intel Realsense D455, downward-facing mvBluefox
    global shutter color camera, and Pixracer flight controller.}
\end{figure}

The proposed framework is deployed to the aerial system of dimensions
\SI{0.6}{\meter} $\times$ \SI{0.3}{\meter} $\times$ \SI{0.3}{\meter} and
mass~\SI{2.5}{\kilo\gram}~(\cref{fig:omicron01}). This design is an improved
version of the robot from~\citep{tabib_autonomous_2021} with a higher motor
constant, longer flight time, and larger depth sensing range (Intel Realsense
D455). The onboard companion computers, state estimation system, the
control system remain the same.

The parameters common to all the experiments are listed in~\cref{tab:params}.
These parameters remain constant throughout teleoperation.
Note that the operator does not need to specify a maximum velocity
parameter found in most multirotor motion planning frameworks. Instead, the maximum velocity
is determined in each planning round based on the local map extents (\cref{ssec:mp_generation}).
\edit{Note that the $\delres$ parameter specifies the change in map resolution after each collision checking
iteration. Thus, this parameter should be specified based on the desired planning rate and the
available onboard compute.}
\begin{table}
\begin{center}
    \begin{tabular}{|c|c|c|c|}
        \hline
        Parameter & Value & Parameter & Value\\
        \hline
        $\plantime$ & \SI{0.1}{\second} & $\yvoxels$ & $20$\\
        $\maptime$ & \SI{0.08}{\second} & $\zvoxels$ & $20$\\
        $\sensetime$ & \SI{0.07}{\second} & $\sensorrange$ & \SI{10.0}{\metre}\\
        $\delres$ & \SI{0.01}{\metre} & $\robotradius$ & \SI{0.3}{\metre}\\
        $\xvoxels$ & $40$ & $\collisionradius$ & \SI{0.1}{\metre}\\
        \hline
    \end{tabular}
    \caption{\label{tab:params}Parameters common to all experiments.}
\end{center}
\end{table}

%% file: content/results.tex
The adaptive teleoperation framework with hierarchical collision avoidance is
implemented with a single thread on a CPU in C++ with the Robot Operating System
(ROS) middleware. A \SI{3.7}{\giga\hertz} Intel Core i9-10900K CPU with $32$ GB RAM is used for
the simulation experiments. A \SI{1.8}{\giga\hertz} Intel Core i7-8550U CPU with $32$ GB RAM is
used for the hardware experiments. We evaluate the approach with four
teleoperation scenarios, two for simulation experiments (\emph{Window Scenario},
\emph{Varying-Clutter Cave Scenario}) and two for hardware experiments
(\emph{Door Scenario}, \emph{Cave Scenario}).  Experimentation for the Cave
Scenario occurred at a cave on the Barbara Schomer Cave Preserve in Clarion
County, PA.

\subsection{Simulation Experiments}

\begin{figure*}
    \centering
    \begin{minipage}{3cm}
      \vspace{-4.6cm}
    \ifthenelse{\equal{\arxivmode}{true}}
    {\subfloat[\label{sfig:window_scenario}Window scenario]{\includegraphics[width=\linewidth,trim=0 0 0 0,clip]{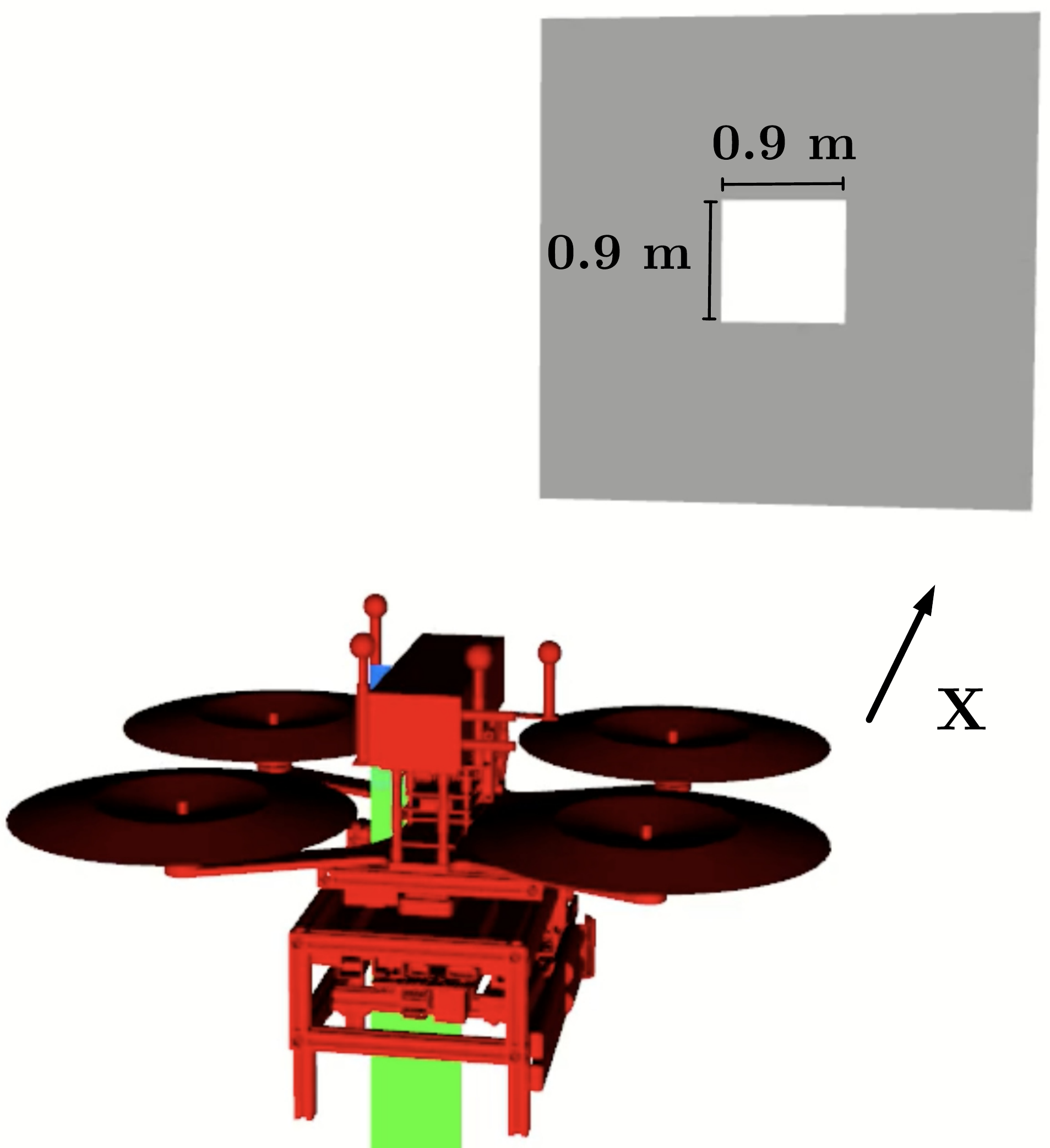}}}
    {\subfloat[\label{sfig:window_scenario}Window scenario]{\includegraphics[width=\linewidth,trim=0 0 0 0,clip]{figures/window_0_45m_scenario.eps}}}\\
      \subfloat[\label{sfig:joystick}Joystick]{\includegraphics[width=\linewidth,trim=0 0 0 0,clip]{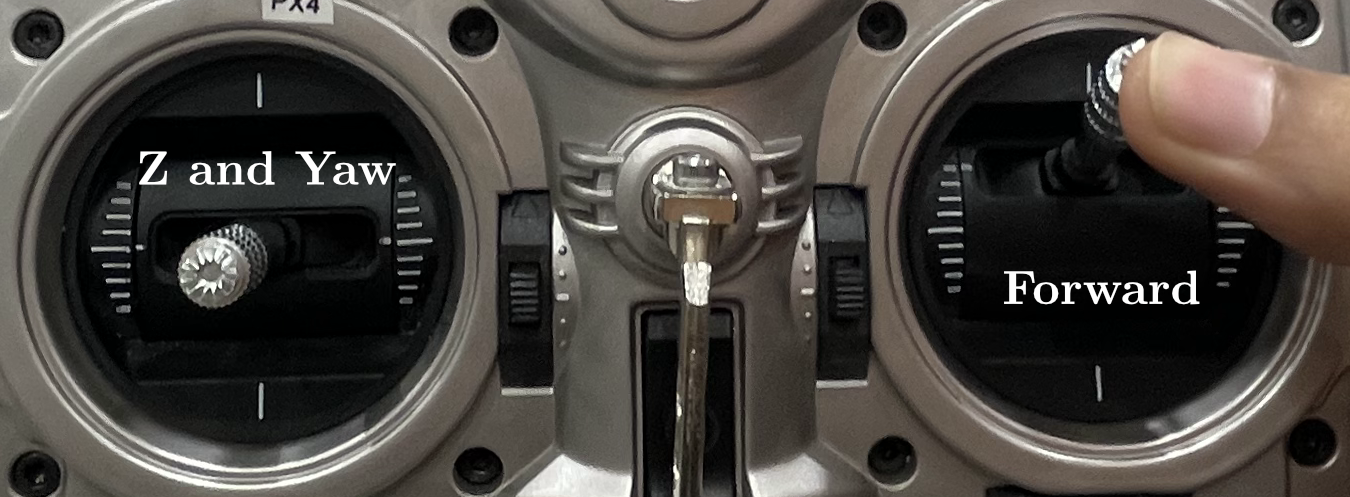}}
    \end{minipage}%
    \ifthenelse{\equal{\arxivmode}{true}}
    {\subfloat[\label{sfig:speed}Linear speed over distance]{\includegraphics{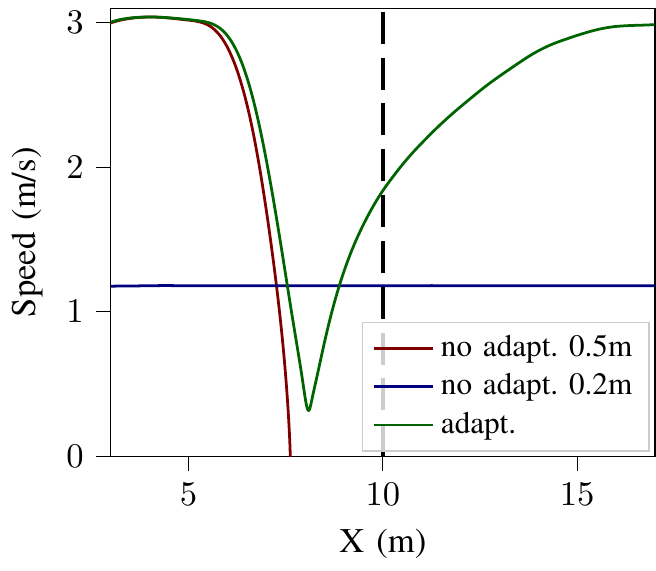}}}
    {\subfloat[\label{sfig:speed}Linear speed over distance]{\input{figures/speed.tex}}}
    \ifthenelse{\equal{\arxivmode}{true}}
    {\subfloat[\label{sfig:map_res}Voxel size over distance]{\includegraphics{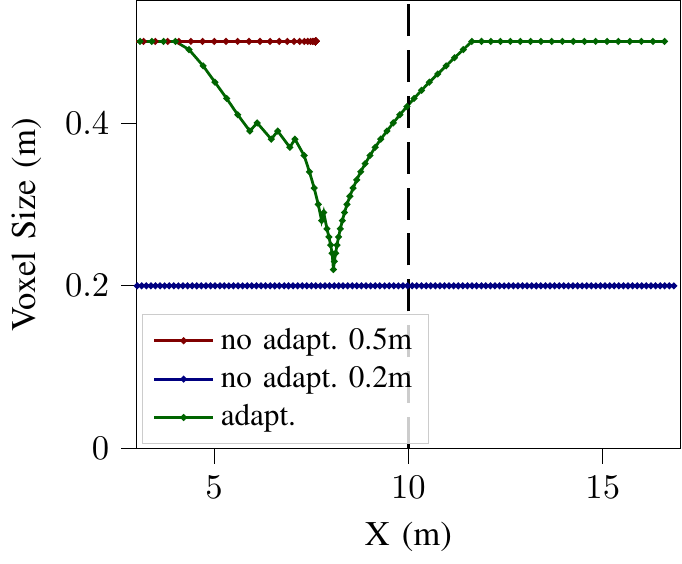}}}
    {\subfloat[\label{sfig:map_res}Voxel size over distance]{\input{figures/map_res.tex}}}
    \caption{\label{fig:window_test}Performance comparison for the simulated
    window teleoperation task.~\protect\subref{sfig:window_scenario} depicts the initial
    conditions for the task. A multirotor hovers at a distance of
    \SI{10}{\metre} from a window of dimensions $\SI{0.9}{\metre} \times
    \SI{0.9}{\metre}$. The operator controls the multirotor via the joystick
    shown in~\protect\subref{sfig:joystick}. The operator intends to go forward at the
    highest speed possible.~\protect\subref{sfig:speed} and \protect\subref{sfig:map_res} show the variation of
    the forward speed and the local map voxel size as a function of the distance from
    the window for the three teleoperation approaches.}
\end{figure*}

\emph{Window Scenario}: A multirotor is placed $\SI{10}{\metre}$ away from a
simulated window of dimensions $\SI{0.9}{\metre} \times
\SI{0.9}{\metre}$~(\cref{sfig:window_scenario}). The operator intends to fly
the multirotor in the forward direction, through the window, at the maximum
possible speed~(\cref{sfig:joystick}). The teleoperation task is successful if
the multirotor passes through the window without collisions and without the
operator having to lower the raw joystick input.
Three teleoperation methodologies are compared: (1) the proposed adaptive
approach with $\maxres = \SI{0.5}{\metre}$ and $\minres = \SI{0.1}{\metre}$; (2)
a non-adaptive approach with a fixed voxel size, $\mapres_1 = \SI{0.2}{\metre}$;
and (3) a non-adaptive approach with a fixed voxel size, $\mapres_2 =
\SI{0.5}{\metre}$.  These parameters are chosen such that for $\mapres_1 =
\SI{0.2}{\metre}$ the local map is fine enough for the window to be visible
while $\mapres_2 = \SI{0.5}{\metre}$ leads to a local map that is too coarse for
it.

The results for the \emph{Window Scenario} are shown in~\cref{fig:window_test}.
We plot the speed and the local map voxel size over the $X$ coordinate with
respect to the window ($X = 0$) in~\cref{sfig:speed,sfig:map_res} respectively.
For the non-adaptive case with $\mapres_1 = \SI{0.2}{\metre}$ voxel size, the
operator teleoperates the multirotor through the window without collision at a
constant speed of $\SI[per-mode=symbol]{1.17}{\metre\per\second}$.  For the
non-adaptive case with $\mapres_2 = \SI{0.5}{\metre}$, the operator is not able
to teleoperate the multirotor through the window while achieving a top speed of
$\SI[per-mode=symbol]{3.03}{\metre\per\second}$ in open space. For the adaptive
case, the operator can teleoperate through the window while being able
to achieve a top speed of $\SI[per-mode=symbol]{3.03}{\metre\per\second}$ in
open space, automatically slowing down to move through the window, and attaining
the same top speed after passing through the window. The automatic slow down is
expected due to the adaptation in the local map voxel size from $\maxres =
\SI{0.5}{\metre}$ down to $\SI{0.25}{\metre}$. Thus, the proposed approach
allows speed modulation without requiring the operator to adjust the joystick
input. Note that the speed attained close to the window in the adaptive
case is lower than the constant speed of \SI{1.17}{\metre\per\second} achieved
in the non-adaptive \SI{0.2}{\metre} voxel size case. This is because the length
of the motion primitive when the robot is decelerating is longer than the case
when the robot is moving at a constant speed. The collision checker marks the
longer motion primitive infeasible, which results in a speed modulation that
decelerates the robot to lower than \SI{1.17}{\metre\per\second}.
This behavior may be changed through a collision checker that utilizes the higher
derivatives of position, but this is left as future work.

\begin{figure*}
    \begin{minipage}{0.6\textwidth}
    \centering
    \subfloat[\label{sfig:cave-sim-env}Simulated cave with varying clutter]{\includegraphics[width=\linewidth]{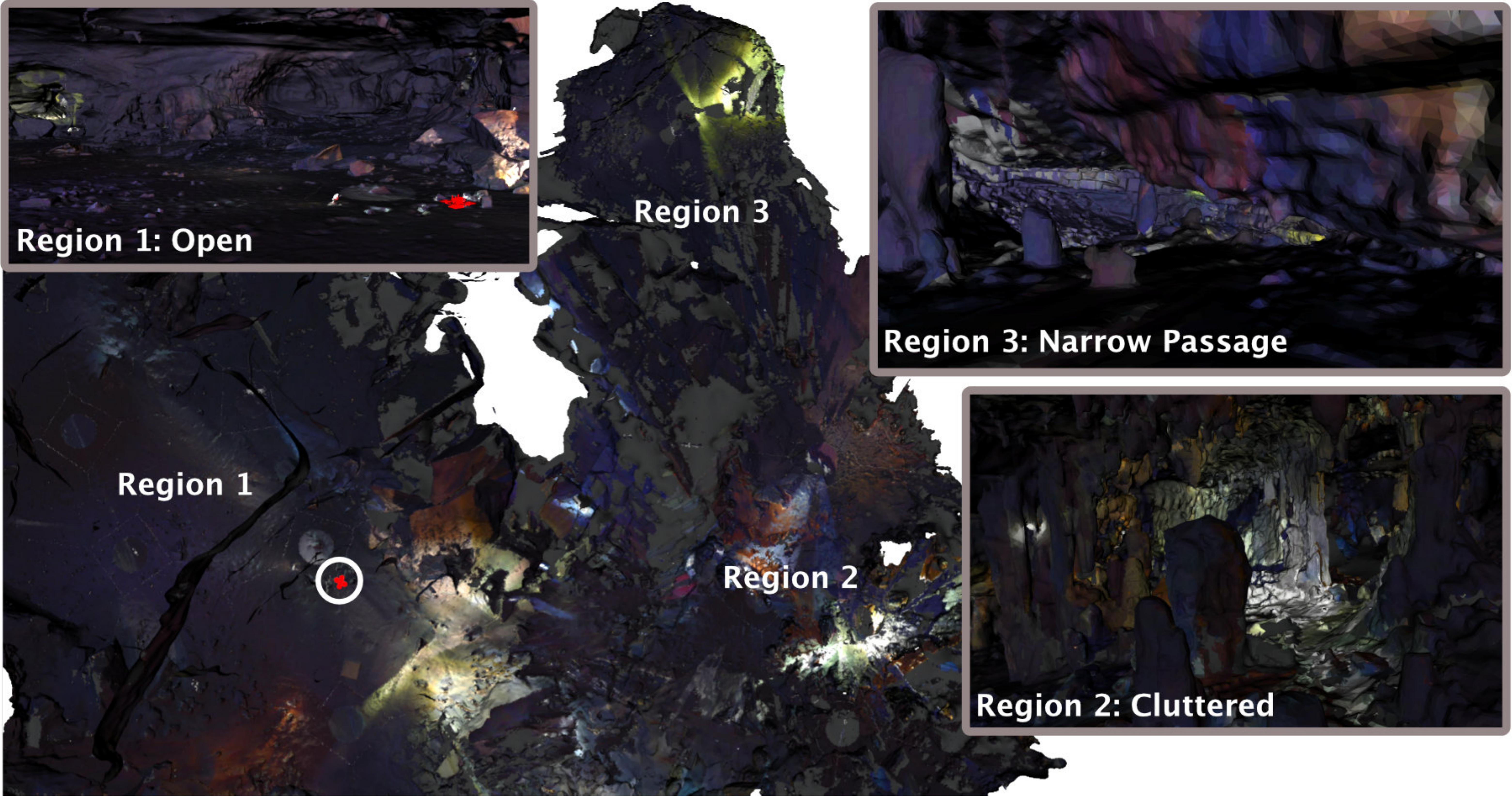}}
    \end{minipage}%
    \begin{minipage}{0.4\textwidth}
    \centering
    \subfloat[\label{sfig:adapt-heatmap}Speed heatmap for adaptive method]{\includegraphics[width=\linewidth]{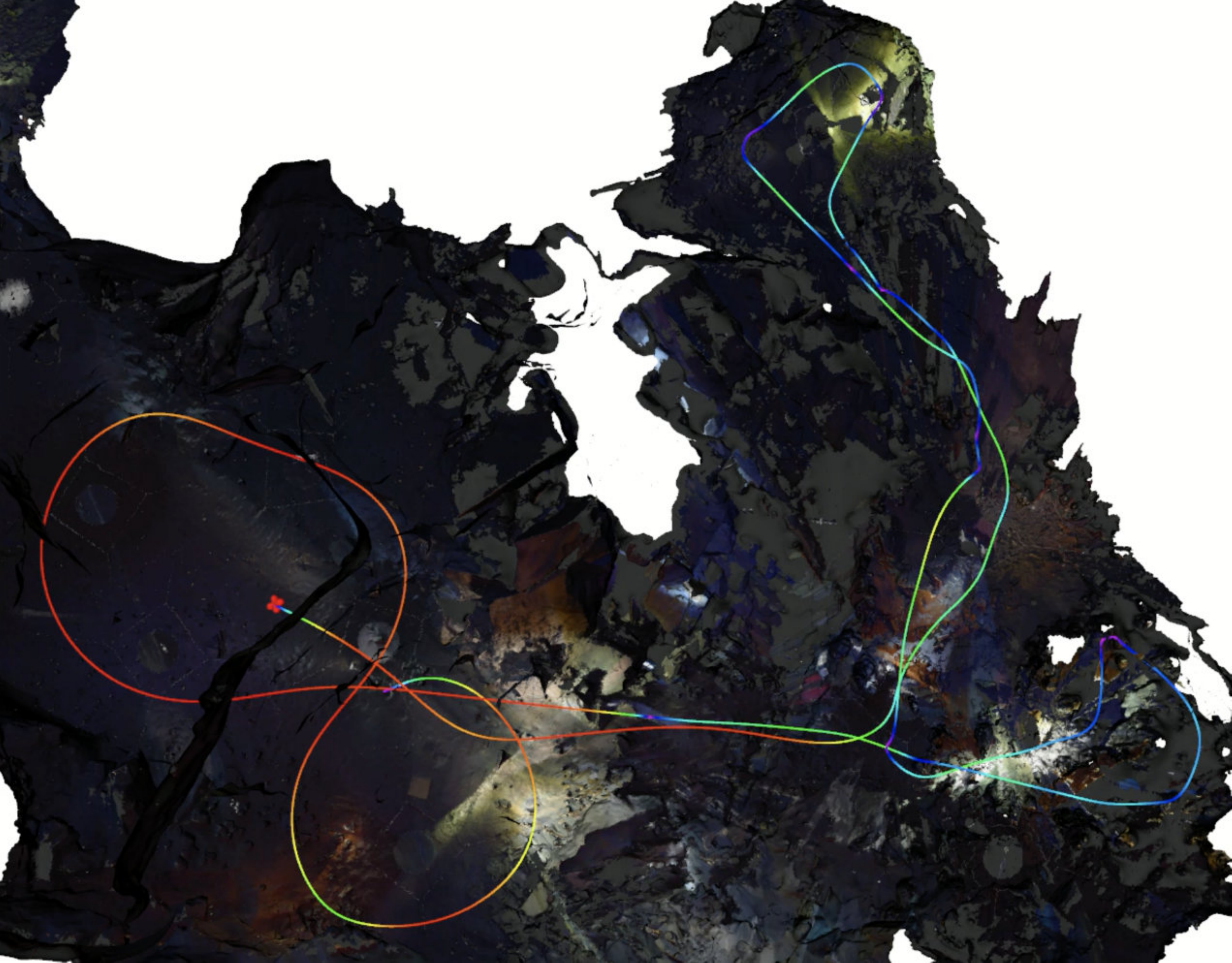}}
    \end{minipage}\\
    \begin{minipage}{\textwidth}
    \centering
    \ifthenelse{\equal{\arxivmode}{true}}
    {\subfloat[\label{sfig:cave-sim-noadapt-1}No Adaptation \SI{0.5}{\metre}]{\includegraphics{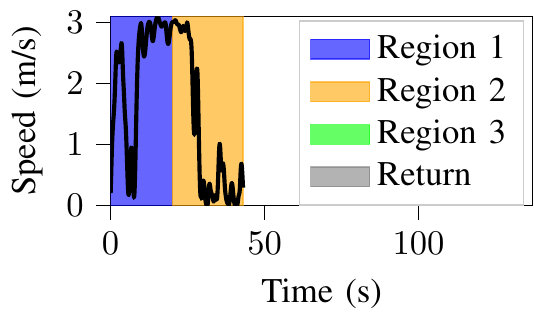}}}
    {\subfloat[\label{sfig:cave-sim-noadapt-1}No Adaptation \SI{0.5}{\metre}]{\input{figures/cave-sim-no-adapt-0-5m.tex}}}%
    \ifthenelse{\equal{\arxivmode}{true}}
    {\subfloat[\label{sfig:cave-sim-noadapt-2}No Adaptation \SI{0.2}{\metre}]{\includegraphics{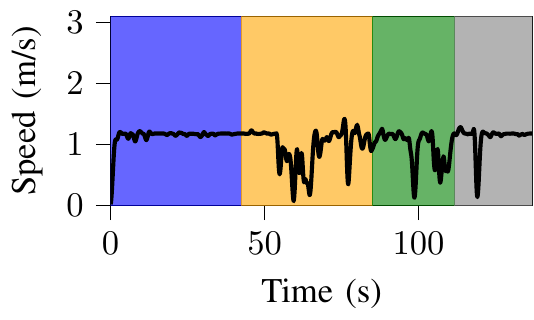}}}
    {\subfloat[\label{sfig:cave-sim-noadapt-2}No Adaptation \SI{0.2}{\metre}]{\input{figures/cave-sim-no-adapt-0-2m.tex}}}%
    \ifthenelse{\equal{\arxivmode}{true}}
    {\subfloat[\label{sfig:cave-sim-adapt}Adaptation]{\includegraphics{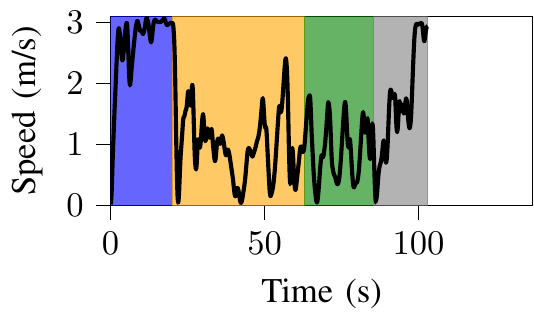}}}
    {\subfloat[\label{sfig:cave-sim-adapt}Adaptation]{\input{figures/cave-sim-adapt.tex}}}
    \end{minipage}\\
    \begin{minipage}{\textwidth}
    \centering
    \subfloat[\label{stab:cave-sim-tab}Comparison of teleoperation statistics]{\input{figures/cave-sim-table.tex}}
    \end{minipage}
    \caption{\label{fig:varying-clutter-cave}
    Performance comparison for the\wennie{~\protect\subref{sfig:cave-sim-env}} varying-clutter cave
    scenario with three different spaces:
    \wennie{R}egion 1 is an open space, \wennie{R}egion 2 is cluttered, and \wennie{R}egion 3 is a narrow
    passage. The speeds achieved by the robot for each method are plotted over time
    in~\protect\subref{sfig:cave-sim-noadapt-1},~\protect\subref{sfig:cave-sim-noadapt-2},
    and~\protect\subref{sfig:cave-sim-adapt}.
    \wennie{The} graphs and the table in~\protect\subref{stab:cave-sim-tab} \wennie{demonstrate}
    that the\wennie{~\protect\subref{sfig:cave-sim-noadapt-1} No Adaptation~\SI{0.5}{\meter} variant
      cannot complete} the circuit, while both the\wennie{~\protect\subref{sfig:cave-sim-noadapt-2}
      No Adaptation~\SI{0.2}{\meter} and~\protect\subref{sfig:cave-sim-adapt} Adaptation variants}
    successfully \wennie{traverse} all regions. Our method completes the circuit in \wennie{the least}
    time while modulating speeds, as \wennie{illustrated in} the heatmap in~\protect\subref{sfig:adapt-heatmap}.
    A video of this experiment can be found at \url{https://youtu.be/VjyoPVXT8WY}.
    }
\end{figure*}

\emph{Varying-Clutter Cave Scenario:} A multirotor is placed in a simulated cave
environment containing varying amounts of clutter~(\cref{sfig:cave-sim-env}).
Region 1 is an open space\wennie{, so the expected operation is high-speed multirotor teleoperation.}
Region 2 is a cluttered space where \wennie{speeds must be reduced} to
ensure safety.  Region 3 contains a narrow entrance to a larger passage that must be visible in the
local occupancy map to enable robot access. The teleoperation
task is to fly from Region 1 to Regions 2 and 3 and return to Region 1
(\cref{sfig:adapt-heatmap}). The operator supplies the directional inputs from
the joystick but the forward speed input is always the maximum value. The three
teleoperation methods from the \emph{Window Scenario} are used here without
modification in parameters. The methods are compared based on: (1) whether they
allow access to all three regions and (2) the total time taken for the
teleoperation task.~\Cref{sfig:cave-sim-noadapt-1,sfig:cave-sim-noadapt-2,sfig:cave-sim-adapt}
show the variation of the speed over time for each of the methods. Without
adaptation, the method with a high voxel size can traverse Region 1 at
high speeds and can enter Region 2 partially. However, it is unable to
access the other regions. When the voxel size is lower or adaptive, the operator
can teleoperate through all regions and return to Region 1. However,
without adaptation, the speeds are lower compared to the
adaptive-speed case. Consequently, the total time taken for the
non-adaptive \SI{0.2}{\metre} case is \SI{137}{\second} versus \SI{103}{\second}
for the adaptive-speed case.~\Cref{stab:cave-sim-tab} shows the distribution of
the time taken with the different phases of the teleoperation task. The
adaptive-speed method achieves lower times and higher average speeds in most
phases. These results demonstrate the efficacy of the proposed method in
environments with varying amounts of clutter.

\subsection{Real-World Experiments}

\begin{figure}
    \centering
    \begin{minipage}{0.5\textwidth}
        \subfloat[\label{sfig:door_scenario}Door scenario]{\includegraphics[height=2.9cm,trim=0 0 0 0,clip]{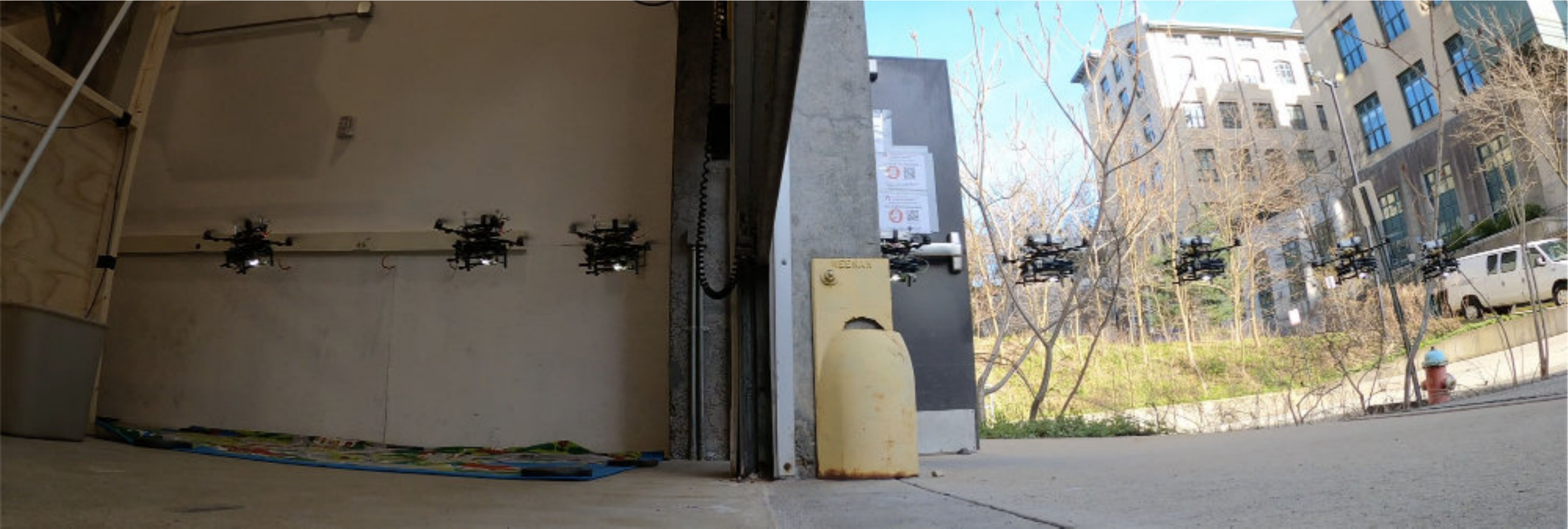}}
    \end{minipage}\\
    \begin{minipage}{0.5\linewidth}
    \ifthenelse{\equal{\arxivmode}{true}}
    {\subfloat[\label{sfig:speed_hw}Linear speed over distance]{\includegraphics{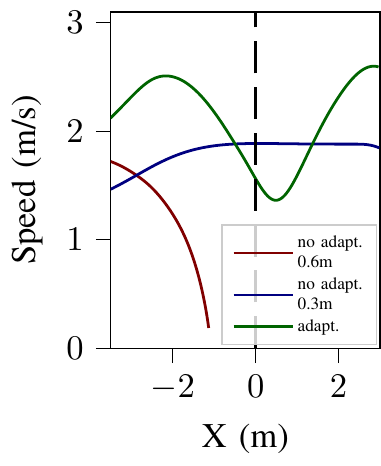}}}
    {\subfloat[\label{sfig:speed_hw}Linear speed over distance]{\input{figures/hw_speed.tex}}}
    \end{minipage}%
    \begin{minipage}{0.5\linewidth}
    \ifthenelse{\equal{\arxivmode}{true}}
    {\subfloat[\label{sfig:map_res_hw}Voxel size over distance]{\includegraphics{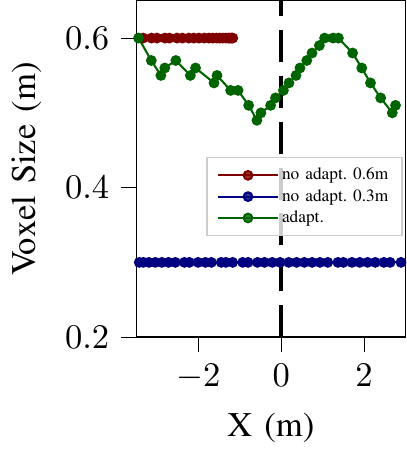}}}
    {\subfloat[\label{sfig:map_res_hw}Voxel size over distance]{\input{figures/hw_map_res.tex}}}
    \end{minipage}
    \caption{\label{fig:door_test}Performance comparison for the door
    teleoperation task.~\wennie{\protect\subref{sfig:door_scenario} a} robot starts at hover from
    outside a building and the operator intends to enter the building at the
    maximum possible forward speed (\cref{sfig:joystick}) through a door of
    width $\SI{0.9}{\metre}$.~\wennie{\protect\subref{sfig:speed_hw} and~\protect\subref{sfig:map_res_hw}} show the
    speeds and voxel sizes as a function of distance from the door. A video
    of this experiment can be found at \url{https://youtu.be/VjyoPVXT8WY}.}
\end{figure}

\emph{Door Scenario}: A multirotor is hovering outside a building,
about $\SI{12}{\metre}$ away from a door, which has a frame with a
$\SI{0.9}{\metre}$ width~(\cref{sfig:door_scenario}). The operator's
intent and the measures of success are the same as in the \emph{Window
Scenario}, this time flying through the door to enter the
building. The same teleoperation methodologies are compared as in the
\emph{Window Scenario}, with $\maxres = \SI{0.6}{\metre}$, $\minres =
\SI{0.3}{\metre}$, $\mapres_1 = \SI{0.3}{\metre}$, and $\mapres_2 =
\SI{0.6}{\metre}$. The choice of these parameters is motivated by the
same reasons as in the \emph{Window Scenario}.

The results for the \emph{Door Scenario} are shown in~\cref{fig:door_test}. Just
like the \emph{Window Scenario}, the speed and local map voxel sizes are plotted
against the distance from the door ($X = 0$)
in~\cref{sfig:speed_hw,sfig:map_res_hw}, respectively. We observe similar results
as in the \emph{Window Scenario}. For the non-adaptive case with $\mapres_1 =
\SI{0.3}{\metre}$, the operator can teleoperate through the door at a
constant speed of $\SI[per-mode=symbol]{1.88}{\metre\per\second}$. For the
non-adaptive case with $\mapres_1 = \SI{0.6}{\metre}$, the operator is not able
to teleoperate through the door while achieving high speeds outside the
building. For the adaptive case, the operator can teleoperate
through the door while being able to achieve a maximum speed of
$\SI[per-mode=symbol]{2.50}{\metre\per\second}$ outside the building, automatically
slowing down at the door and achieving the same maximum speed again after entering the building.
Thus, the \emph{Door Scenario} experiment demonstrates the efficacy of the proposed teleoperation
method on the computationally-constrained multirotor system shown in~\cref{fig:omicron01}.


\begin{figure}
    \centering
    \begin{minipage}{0.5\textwidth}
    \ifthenelse{\equal{\arxivmode}{true}}
    {\subfloat[\label{sfig:cave_scenario}Cave scenario]{\includegraphics[height=2.9cm,trim=200 20 10 70,clip]{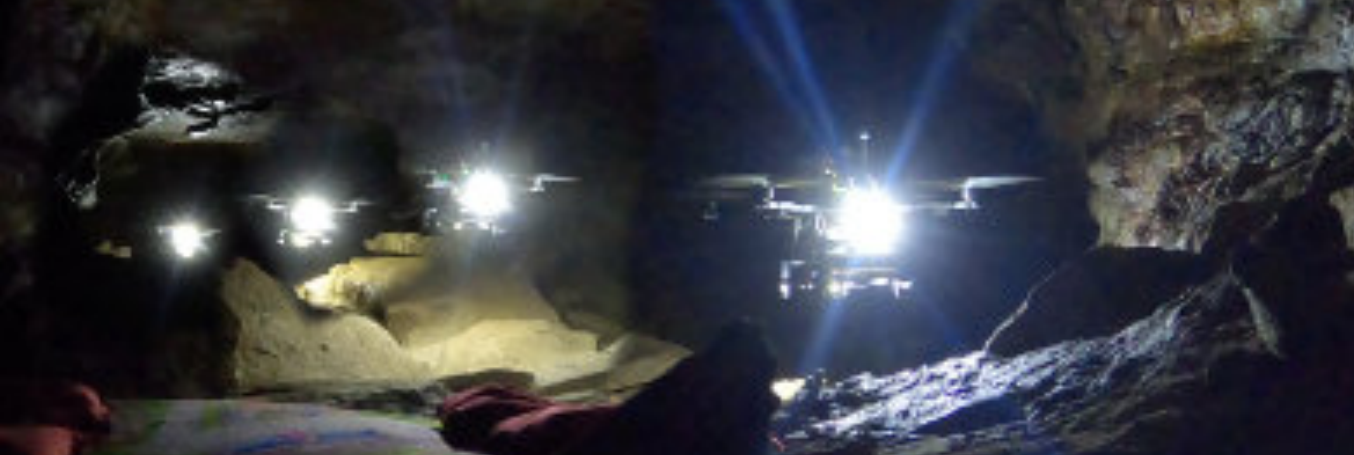}}}
    {\subfloat[\label{sfig:cave_scenario}Cave scenario]{\includegraphics[height=2.9cm,trim=200 20 10 70,clip]{./images/cave_composite_flattened-compressed.eps}}}
    \end{minipage}\\
    \begin{minipage}{0.25\textwidth}
    \ifthenelse{\equal{\arxivmode}{true}}
    {\subfloat[\label{sfig:speed_hw_cave}Linear speed over time]{\includegraphics{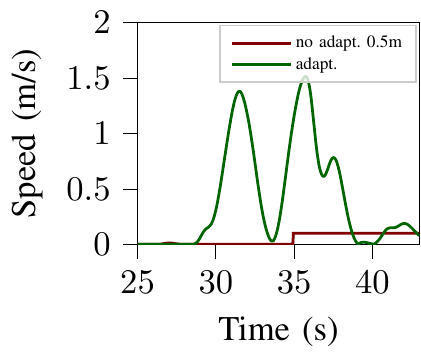}}}
    {\subfloat[\label{sfig:speed_hw_cave}Linear speed over time]{\input{figures/hw_speed_cave.tex}}}
    \end{minipage}%
    \begin{minipage}{0.25\textwidth}
    \ifthenelse{\equal{\arxivmode}{true}}
    {\subfloat[\label{sfig:map_res_hw_cave}Voxel size over time]{\includegraphics{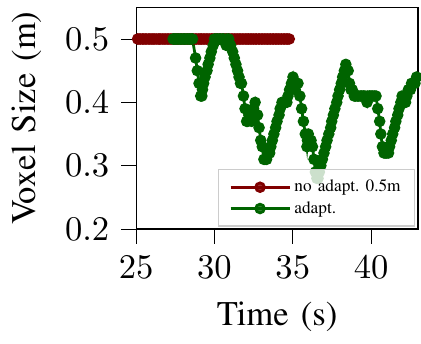}}}
    {\subfloat[\label{sfig:map_res_hw_cave}Voxel size over time]{\input{figures/hw_map_res_cave.tex}}}
    \end{minipage}
    \caption{\label{fig:cave_test}Performance comparison for the cave
    teleoperation task.~\wennie{\protect\subref{sfig:cave_scenario} a} robot starts at hover from
    a relatively spacious part of a cave passage. The operator intends to go
    through the passage at the maximum possible forward speed
    (\cref{sfig:joystick}).~\wennie{\protect\subref{sfig:speed_hw_cave} and \protect\subref{sfig:map_res_hw_cave}} show the speeds
    and map voxel sizes as a function of time. Without map adaptation, the robot is unable to
    go through the narrow passage and the operator must land the robot around the
    \SI{35}{\second} mark. With map adaptation, the speeds are adapted according to the
    environment complexity and the robot traverses the narrow passage. A video of this experiment can be
    found at \url{https://youtu.be/VjyoPVXT8WY}.}
\end{figure}

\emph{Cave Scenario}: A multirotor hover\wennie{s} in a narrow passage
inside of the wild cave shown in~\cref{sfig:cave_entrance2}. The
operator intends to fly through the narrow passage without
substantially altering the raw joystick input. Two teleoperation
methodologies are compared in this case: (1) the adaptive
approach with $\maxres = \SI{0.5}{\metre},$ $\minres = \SI{0.1}{\metre}$ and
(2) the non-adaptive approach with $\mapres = \SI{0.5}{\metre}$.

The results for the \emph{Cave Scenario} are shown in~\cref{fig:cave_test}. In the no-adaptation case with
$\mapres = \SI{0.5}{\metre}$, the operator is not able to teleoperate the multirotor through the narrow passage
of the cave due to the coarse local map. However, for the adaptation case, the voxel size is reduced automatically
to allow for a finer local map for flight through the narrow passage. Thus, the proposed approach allows
teleoperation through a narrow passage where it is difficult to guess the required voxel size of the local map
before starting teleoperation.

\Cref{tab:plan_times} illustrates the mean and standard deviation for total planning
times for the real-world experiments.  As noted in~\cref{ssec:hier_coll_avd},
most of the time complexity of the proposed approach is due to local map
generation. Imposing an upper bound on the number of local map generation steps
per planning round enables us to contain that time complexity and achieve a
consistent planning rate (i.e., the low standard deviation in planning time).

\begin{table}
\begin{center}
    \begin{tabular}{ccc}
        \toprule
        Method & \emph{Door Scenario} & \emph{Cave Scenario}\\
        \midrule
        No adapt. low vox. size & \SI{0.06}{\second} $\pm$ \SI{0.01}{\second} & -- \\
        No adapt. high vox. size & \SI{0.03}{\second} $\pm$ \SI{0.01}{\second} & \SI{0.04}{\second} $\pm$ \SI{0.01}{\second}\\
        Adaptation & \SI{0.06}{\second} $\pm$ \SI{0.03}{\second} & \SI{0.06}{\second} $\pm$ \SI{0.03}{\second}\\
        \bottomrule
    \end{tabular}
    \caption{\label{tab:plan_times}Planning times for the real-world experiments.}
\end{center}
\end{table}

%% file: figures/map_res.tex
\begin{tikzpicture}

\definecolor{darkgray176}{RGB}{176,176,176}
\definecolor{darkorange25512714}{RGB}{255,127,14}
\definecolor{forestgreen4416044}{RGB}{44,160,44}
\definecolor{lightgray204}{RGB}{204,204,204}
\definecolor{steelblue31119180}{RGB}{31,119,180}
\definecolor{maroon}{RGB}{128,0,0}
\definecolor{navy}{RGB}{0,0,128}
\definecolor{darkgreen}{RGB}{0,100,0}

\begin{axis}[
width=0.4\linewidth,
legend cell align={left},
legend style={
  nodes={scale=0.9, transform shape},
  fill opacity=0.8,
  draw opacity=1,
  text opacity=1,
  at={(0.01,0.01)},
  anchor=south west,
  draw=lightgray204
},
tick align=outside,
tick pos=left,
x grid style={darkgray176},
xlabel={X (m)},
xmin=3, xmax=17,
xtick style={color=black},
y grid style={darkgray176},
ylabel={Voxel Size (m)},
ymin=0, ymax=0.55,
ytick style={color=black}
]
\addplot [thick, maroon, mark=*, mark size=0.5, mark options={solid}]
table {%
2.90496878847548 0.5
3.19012846579706 0.5
3.49232529645535 0.5
3.81088074715828 0.5
4.09955180807841 0.5
4.40630866003713 0.5
4.7063628327769 0.5
5.0085638647018 0.5
5.30972184676136 0.5
5.60850977265969 0.5
5.9011481038334 0.5
6.18438120313168 0.5
6.44323687480729 0.5
6.6795862909215 0.5
6.88604909479375 0.5
7.06104646286208 0.5
7.20597333684512 0.5
7.32348816736878 0.5
7.41219392959896 0.5
7.4838877620433 0.5
7.53605510561631 0.5
7.57184644663061 0.5
7.59471172432412 0.5
7.60810729119223 0.5
7.61514585370307 0.5
7.61833552286471 0.5
7.61948648343044 0.5
7.6197522289966 0.5
7.61975214752225 0.5
7.61972921237179 0.5
7.61971505380116 0.5
7.61967018178006 0.5
7.61958232328633 0.5
7.61950483343598 0.5
7.61950786018302 0.5
7.6196072926962 0.5
7.61976559821391 0.5
7.61992890081538 0.5
7.62005782383088 0.5
7.62013505074066 0.5
7.6201592165763 0.5
7.62014127580225 0.5
7.62009925047174 0.5
7.62005821616533 0.5
7.62003691103946 0.5
7.6200424094103 0.5
7.62007297483792 0.5
7.62012566885588 0.5
7.62018071153605 0.5
7.62023495960565 0.5
7.62027673679902 0.5
7.62030087661638 0.5
7.62030854794108 0.5
7.62030270405625 0.5
7.62028935934688 0.5
7.62027412503027 0.5
7.62026080260696 0.5
7.62025197034901 0.5
7.62024851593679 0.5
7.62024964266254 0.5
7.62025346912123 0.5
7.62025763383118 0.5
7.62026113206461 0.5
7.62026376182273 0.5
7.62026476674167 0.5
7.62026461356096 0.5
7.62026466964085 0.5
7.62026469529931 0.5
7.62026544949231 0.5
7.62026645433394 0.5
7.62026763095842 0.5
7.62026938556587 0.5
7.62027258215266 0.5
7.62027717229373 0.5
7.62028224496379 0.5
7.62028580035187 0.5
7.62028752022979 0.5
7.62028819456351 0.5
7.62028752161046 0.5
7.62028596964334 0.5
7.62028382814733 0.5
7.62028133012345 0.5
7.62027899047701 0.5
7.62027670557168 0.5
7.6202754741556 0.5
7.62027636278252 0.5
7.62027873884045 0.5
7.62028254049278 0.5
7.62028638841632 0.5
7.62028917372375 0.5
7.62029088755717 0.5
7.62029106754604 0.5
7.62029061320888 0.5
7.62028948497619 0.5
7.62028776450834 0.5
7.62028661255284 0.5
7.6202872806772 0.5
7.62028913843788 0.5
7.62029301815757 0.5
7.62029625268056 0.5
7.62029869909825 0.5
7.62030038831823 0.5
7.62030123106517 0.5
7.62030003319464 0.5
7.62029735334383 0.5
7.62029420046301 0.5
7.62029117797235 0.5
7.62028954064463 0.5
7.62029048288429 0.5
7.62029478028388 0.5
7.62030195384667 0.5
7.62031085347745 0.5
7.62032034872098 0.5
7.62032846440597 0.5
7.62033415145471 0.5
};
\addlegendentry{no adapt. 0.5m}
\addplot [thick, navy, mark=*, mark size=0.5, mark options={solid}]
table {%
3.02428833545049 0.200000002980232
3.14201070884875 0.200000002980232
3.26094038299759 0.200000002980232
3.37872573022465 0.200000002980232
3.49536421048085 0.200000002980232
3.61321070426958 0.200000002980232
3.7310907137101 0.200000002980232
3.84901120524385 0.200000002980232
3.96697710729641 0.200000002980232
4.08498724146824 0.200000002980232
4.20303632662139 0.200000002980232
4.32111370895846 0.200000002980232
4.43920491579359 0.200000002980232
4.5572926490437 0.200000002980232
4.67536215650031 0.200000002980232
4.79340181782321 0.200000002980232
4.91140979359848 0.200000002980232
5.02939084135036 0.200000002980232
5.15324964821862 0.200000002980232
5.26532657018481 0.200000002980232
5.38334217009449 0.200000002980232
5.5013808955419 0.200000002980232
5.61941382713993 0.200000002980232
5.73742462606422 0.200000002980232
5.85540915291599 0.200000002980232
5.97337412859796 0.200000002980232
6.09250977114981 0.200000002980232
6.2104713651189 0.200000002980232
6.32726690421321 0.200000002980232
6.44525972025619 0.200000002980232
6.56327030911321 0.200000002980232
6.68129390115019 0.200000002980232
6.80639658273755 0.200000002980232
6.9173209619341 0.200000002980232
7.03530666733828 0.200000002980232
7.15327448635983 0.200000002980232
7.27123541167361 0.200000002980232
7.38920302690907 0.200000002980232
7.50718601540028 0.200000002980232
7.61928320069616 0.200000002980232
7.73728624515216 0.200000002980232
7.85528880026886 0.200000002980232
7.97328823837123 0.200000002980232
8.09128429255888 0.200000002980232
8.20927826518488 0.200000002980232
8.32137500896586 0.200000002980232
8.44055822331828 0.200000002980232
8.55738643019067 0.200000002980232
8.67540072040307 0.200000002980232
8.79341081251149 0.200000002980232
8.9114092006857 0.200000002980232
9.02939506715618 0.200000002980232
9.14736654977354 0.200000002980232
9.2653281684589 0.200000002980232
9.3891861221391 0.200000002980232
9.50715712636133 0.200000002980232
9.62514728056322 0.200000002980232
9.74315803181469 0.200000002980232
9.86118590204357 0.200000002980232
9.9804053558327 0.200000002980232
10.0972657375204 0.200000002980232
10.2152950483166 0.200000002980232
10.339202894011 0.200000002980232
10.4571892809007 0.200000002980232
10.5751649663124 0.200000002980232
10.6931419399696 0.200000002980232
10.8111227718925 0.200000002980232
10.9291118391653 0.200000002980232
11.0471145129879 0.200000002980232
11.1651524424281 0.200000002980232
11.2832184817676 0.200000002980232
11.4012761465514 0.200000002980232
11.5204717536904 0.200000002980232
11.6372623612495 0.200000002980232
11.7552014898701 0.200000002980232
11.8802011372746 0.200000002980232
11.9910566991734 0.200000002980232
12.1102086675077 0.200000002980232
12.2270577653956 0.200000002980232
12.3451120965433 0.200000002980232
12.4631526882302 0.200000002980232
12.5811604956916 0.200000002980232
12.6991336256984 0.200000002980232
12.817088655602 0.200000002980232
12.9362273677125 0.200000002980232
13.0530221679004 0.200000002980232
13.1710137027682 0.200000002980232
13.2890227268051 0.200000002980232
13.4070438337526 0.200000002980232
13.5250666463577 0.200000002980232
13.6430749937944 0.200000002980232
13.7610791641284 0.200000002980232
13.8790973084513 0.200000002980232
13.997122069762 0.200000002980232
14.1151340502432 0.200000002980232
14.2331226079041 0.200000002980232
14.3510918018488 0.200000002980232
14.4690498761892 0.200000002980232
14.5870078522766 0.200000002980232
14.7049717612149 0.200000002980232
14.8229471365738 0.200000002980232
14.9409421817003 0.200000002980232
15.0601433990833 0.200000002980232
15.176999419072 0.200000002980232
15.2950289621762 0.200000002980232
15.4130321190988 0.200000002980232
15.5310079765006 0.200000002980232
15.6489644738003 0.200000002980232
15.7669145962125 0.200000002980232
15.8860465505447 0.200000002980232
16.0028307411688 0.200000002980232
16.1208157582228 0.200000002980232
16.2388189553186 0.200000002980232
16.356831779252 0.200000002980232
16.4760303763933 0.200000002980232
16.5928684802312 0.200000002980232
16.7120563125656 0.200000002980232
16.8288665028389 0.200000002980232
};
\addlegendentry{no adapt. 0.2m}
\addplot [thick, darkgreen, mark=*, mark size=0.5, mark options={solid}]
table {%
3.09938722765816 0.5
3.40149470021748 0.5
3.70483669162799 0.5
4.0086844292409 0.5
4.35808506397478 0.490000009536743
4.72200174257183 0.470000028610229
5.02450270178604 0.450000047683716
5.32610431897458 0.430000066757202
5.61136519385235 0.410000085830688
5.92293829821554 0.390000104904175
6.11166756067835 0.400000095367432
6.47415452068713 0.380000114440918
6.63884393054244 0.390000104904175
6.95116112655114 0.370000123977661
7.08297351835162 0.380000114440918
7.3184456701269 0.360000133514404
7.4557097004207 0.340000152587891
7.57818247032406 0.320000171661377
7.6815297410401 0.300000190734863
7.76907928225234 0.28000020980835
7.8223905753984 0.290000200271606
7.90640521042944 0.270000219345093
7.95688536167064 0.260000228881836
8.00143180703149 0.250000238418579
8.03915310780409 0.240000233054161
8.07257550804801 0.220000222325325
8.0964556109634 0.230000227689743
8.12894166374358 0.240000233054161
8.16450586546149 0.250000238418579
8.20477911563031 0.260000228881836
8.25040523851768 0.270000219345093
8.30185478468765 0.28000020980835
8.35683180593711 0.290000200271606
8.42212137524004 0.300000190734863
8.49596696752534 0.31000018119812
8.57913502762607 0.320000171661377
8.67217353016576 0.330000162124634
8.77539011146009 0.340000152587891
8.8889075962061 0.350000143051147
9.013961160401 0.360000133514404
9.14643081428934 0.370000123977661
9.28985111724954 0.380000114440918
9.45036217200392 0.390000104904175
9.61223744453002 0.400000095367432
9.78255089248662 0.410000085830688
9.962698940686 0.420000076293945
10.1466587674052 0.430000066757202
10.3394234566166 0.440000057220459
10.5389849727576 0.450000047683716
10.7556090758134 0.460000038146973
10.9705016302512 0.470000028610229
11.1871136030004 0.480000019073486
11.4141741463957 0.490000009536743
11.6308604643547 0.5
11.8812358351416 0.5
12.1206734399614 0.5
12.3679696966146 0.5
12.6232996490149 0.5
12.8788322963232 0.5
13.1419770834812 0.5
13.4100547173285 0.5
13.6858325008701 0.5
13.9608472481253 0.5
14.2427415712852 0.5
14.5280981042141 0.5
14.8163270028948 0.5
15.1101073912149 0.5
15.4004815992346 0.5
15.6958326674281 0.5
15.9926719930775 0.5
16.290347058928 0.5
16.5884344316506 0.5
};
\addlegendentry{adapt.}
\path [draw=black, line width=1pt, dash pattern=on 9.25pt off 4pt]
(axis cs:10,0)
--(axis cs:10,0.55);

\end{axis}

\end{tikzpicture}

%% file: figures/cave-sim-no-adapt-0-5m.tex
\begin{tikzpicture}

\definecolor{darkgray176}{RGB}{176,176,176}
\definecolor{lightgray204}{RGB}{204,204,204}
\definecolor{orange}{RGB}{255,165,0}

\begin{axis}[
width=0.33\textwidth,
height=3.5cm,
legend cell align={left},
legend style={fill opacity=0.8, draw opacity=1, text opacity=1, draw=lightgray204},
tick align=outside,
tick pos=left,
x grid style={darkgray176},
xmin=0, xmax=137,
xtick style={color=black},
y grid style={darkgray176},
ymin=0, ymax=3.1,
ylabel={Speed (m/s)},
xlabel={Time (s)},
ytick style={color=black}
]
\path [draw=blue, fill=blue, opacity=0.6]
(axis cs:0,0)
--(axis cs:0,3.1)
--(axis cs:20,3.1)
--(axis cs:20,0)
--cycle;
\addlegendimage{area legend, draw=blue, fill=blue, opacity=0.6}
\addlegendentry{Region 1}

\path [draw=orange, fill=orange, opacity=0.6]
(axis cs:20,0)
--(axis cs:20,3.1)
--(axis cs:43,3.1)
--(axis cs:43,0)
--cycle;
\addlegendimage{area legend, draw=orange, fill=orange, opacity=0.6}
\addlegendentry{Region 2}

\addlegendimage{area legend, draw=green, fill=green, opacity=0.6}
\addlegendentry{Region 3}

\addlegendimage{area legend, draw=gray, fill=gray, opacity=0.6}
\addlegendentry{Return}

\addplot [very thick, black, forget plot]
table {%
0 0.201622366905212
0.0299999713897705 0.242952942848206
0.059999942779541 0.288204312324524
0.0900000333786011 0.337118864059448
0.125 0.398308515548706
0.164999961853027 0.472615480422974
0.225000023841858 0.589186310768127
0.299999952316284 0.734214305877686
0.345000028610229 0.816184401512146
0.384999990463257 0.884722709655762
0.424999952316284 0.948771595954895
0.460000038146973 1.0009423494339
0.495000004768372 1.04951238632202
0.529999971389771 1.09439861774445
0.565000057220459 1.13582694530487
0.601000070571899 1.17512273788452
0.635999917984009 1.21033334732056
0.674999952316284 1.24646234512329
0.714999914169312 1.28066205978394
0.759999990463257 1.31639409065247
0.815000057220459 1.35728538036346
0.904999971389771 1.42109179496765
1 1.48905336856842
1.05599999427795 1.53165900707245
1.10500001907349 1.57152700424194
1.14999997615814 1.61093008518219
1.19000005722046 1.64870524406433
1.23000001907349 1.68933343887329
1.26999998092651 1.73293077945709
1.30999994277954 1.77947998046875
1.35500001907349 1.835324883461
1.39999997615814 1.89472615718842
1.45000004768372 1.96443569660187
1.50999999046326 2.05196118354797
1.61000001430511 2.20252752304077
1.66999995708466 2.29108858108521
1.71000003814697 2.34640860557556
1.74000000953674 2.38457202911377
1.76499998569489 2.41348075866699
1.78999996185303 2.43925952911377
1.81500005722046 2.46155190467834
1.83500003814697 2.47672176361084
1.85500001907349 2.48946404457092
1.875 2.49977278709412
1.89600002765656 2.50802111625671
1.91499996185303 2.51328754425049
1.93499994277954 2.51669883728027
1.95500004291534 2.51808381080627
1.97500002384186 2.5176305770874
2 2.51478934288025
2.03099989891052 2.50859808921814
2.10999989509583 2.48237991333008
2.1800000667572 2.45425033569336
2.25999999046326 2.42256140708923
2.31500005722046 2.40305733680725
2.36500000953674 2.38750672340393
2.41000008583069 2.37557125091553
2.45499992370605 2.36586809158325
2.5 2.35854506492615
2.54500007629395 2.35354399681091
2.58999991416931 2.35075449943542
2.63499999046326 2.35009860992432
2.6800000667572 2.35145354270935
2.72499990463257 2.35478329658508
2.76999998092651 2.36023592948914
2.81500005722046 2.36789608001709
2.85999989509583 2.37768363952637
2.90499997138977 2.38951349258423
2.95000004768372 2.40343022346497
2.99499988555908 2.41953325271606
3.03999996185303 2.43776488304138
3.08999991416931 2.46035647392273
3.14499998092651 2.48765325546265
3.21499991416931 2.52493357658386
3.39000010490417 2.61975121498108
3.42499995231628 2.63541412353516
3.45499992370605 2.64653968811035
3.48000001907349 2.65370297431946
3.50500011444092 2.65853071212769
3.52500009536743 2.66044306755066
3.54500007629395 2.66040730476379
3.56500005722046 2.65824675559998
3.58500003814697 2.6538200378418
3.60500001907349 2.64702582359314
3.625 2.63781261444092
3.64599990844727 2.6255316734314
3.66499996185303 2.61215591430664
3.68499994277954 2.59582233428955
3.71000003814697 2.57232356071472
3.73499989509583 2.54566955566406
3.76500010490417 2.51003122329712
3.79500007629395 2.47111034393311
3.83599996566772 2.41409373283386
3.91000008583069 2.30634546279907
3.96499991416931 2.22789192199707
4.00500011444092 2.1741418838501
4.04500007629395 2.12416362762451
4.07999992370605 2.08371829986572
4.11999988555908 2.04089140892029
4.17000007629395 1.99105477333069
4.28999996185303 1.87353205680847
4.33599996566772 1.82437312602997
4.38000011444092 1.77424240112305
4.44999980926514 1.69063305854797
4.51100015640259 1.61900901794434
4.55000019073486 1.57626950740814
4.58500003814697 1.54079854488373
4.61999988555908 1.50827217102051
4.65500020980835 1.47850871086121
4.69500017166138 1.44719958305359
4.76000022888184 1.39959025382996
4.82499980926514 1.3512316942215
4.86499977111816 1.3190108537674
4.90500020980835 1.28390657901764
4.92500019073486 1.26509571075439
4.92999982833862 1.25927793979645
4.94000005722046 1.25039684772491
4.97499990463257 1.21405601501465
5.01000022888184 1.1747944355011
5.04500007629395 1.13262259960175
5.08599996566772 1.07966637611389
5.13000011444092 1.01892340183258
5.17999982833862 0.945689916610718
5.2350001335144 0.861155390739441
5.32000017166138 0.725893616676331
5.42500019073486 0.559239983558655
5.48000001907349 0.475751161575317
5.52500009536743 0.411120414733887
5.55999994277954 0.364086985588074
5.59499979019165 0.320736646652222
5.625 0.287157297134399
5.65000009536743 0.2621169090271
5.67500019073486 0.240062594413757
5.69999980926514 0.221220135688782
5.71999979019165 0.208557486534119
5.73999977111816 0.198072791099548
5.76000022888184 0.189755201339722
5.7810001373291 0.183298945426941
5.80000019073486 0.179384589195251
5.82000017166138 0.177124500274658
5.84499979019165 0.176759719848633
5.87099981307983 0.178959369659424
5.89499998092651 0.183025121688843
5.92500019073486 0.190489053726196
5.95499992370605 0.200296401977539
5.9850001335144 0.212277770042419
6.0149998664856 0.226423025131226
6.04500007629395 0.242861270904541
6.07499980926514 0.261825561523438
6.10500001907349 0.283591747283936
6.13500022888184 0.30843448638916
6.16499996185303 0.336596846580505
6.19500017166138 0.368207097053528
6.22499990463257 0.403212666511536
6.25500011444092 0.44136655330658
6.28999996185303 0.489270448684692
6.33599996566772 0.554713726043701
6.34499979019165 0.569619536399841
6.44000005722046 0.710063934326172
6.47499990463257 0.757979393005371
6.50500011444092 0.795949339866638
6.53499984741211 0.830417156219482
6.55999994277954 0.856115937232971
6.58500003814697 0.87882387638092
6.6100001335144 0.898361444473267
6.63000011444092 0.911609292030334
6.65000009536743 0.922674775123596
6.67000007629395 0.931515455245972
6.69000005722046 0.93810772895813
6.71099996566772 0.942605257034302
6.73000001907349 0.94454550743103
6.75 0.944430112838745
6.76999998092651 0.942139148712158
6.78999996185303 0.93771755695343
6.80999994277954 0.931217074394226
6.82999992370605 0.922697305679321
6.85500001907349 0.909311771392822
6.88000011444092 0.893028497695923
6.9060001373291 0.873202800750732
6.92999982833862 0.852463006973267
6.96000003814697 0.8235182762146
6.98999977111816 0.791553258895874
7.02500009536743 0.750943422317505
7.06599998474121 0.699626088142395
7.11499977111816 0.634351849555969
7.18499994277954 0.536916971206665
7.28499984741211 0.398057341575623
7.32999992370605 0.339060187339783
7.36999988555908 0.290150165557861
7.40000009536743 0.256434440612793
7.42999982833862 0.225904583930969
7.45599985122681 0.202462315559387
7.48000001907349 0.183640480041504
7.5 0.170230031013489
7.51999998092651 0.159053683280945
7.53999996185303 0.150257587432861
7.55999994277954 0.143974423408508
7.57499980926514 0.14098048210144
7.59499979019165 0.139353036880493
7.6100001335144 0.139940977096558
7.63000011444092 0.143164277076721
7.65000009536743 0.149179458618164
7.67000007629395 0.157966494560242
7.69000005722046 0.169488310813904
7.71000003814697 0.183694005012512
7.73000001907349 0.200521230697632
7.75 0.219900488853455
7.77500009536743 0.247594475746155
7.80000019073486 0.278972268104553
7.82499980926514 0.313834428787231
7.84999990463257 0.352081775665283
7.88000011444092 0.40227484703064
7.90999984741211 0.456808805465698
7.94500017166138 0.525356769561768
7.98000001907349 0.598657608032227
8.02000045776367 0.687538385391235
8.0649995803833 0.792945384979248
8.11999988555908 0.92765736579895
8.19999980926514 1.12999033927917
8.32999992370605 1.45898580551147
8.39000034332275 1.60477304458618
8.44499969482422 1.73269784450531
8.49600028991699 1.84575366973877
8.54500007629395 1.94885957241058
8.59000015258789 2.03846073150635
8.63500022888184 2.1229133605957
8.67599964141846 2.19516897201538
8.71500015258789 2.25965142250061
8.75500011444092 2.32144713401794
8.79500007629395 2.37890195846558
8.83500003814697 2.43218111991882
8.875 2.48147130012512
8.91499996185303 2.52693963050842
8.94999980926514 2.56368494033813
8.98499965667725 2.59772038459778
9.02499961853027 2.6335711479187
9.0649995803833 2.66654419898987
9.10999965667725 2.70061421394348
9.15499973297119 2.73177003860474
9.19999980926514 2.7602756023407
9.25 2.78922486305237
9.30000019073486 2.8157651424408
9.35499954223633 2.84258341789246
9.40999984741211 2.86713266372681
9.4709997177124 2.89197897911072
9.53499984741211 2.9156858921051
9.5959997177124 2.93594765663147
9.65499973297119 2.95333409309387
9.72000026702881 2.97019672393799
9.76500034332275 2.9798686504364
9.80000019073486 2.98544931411743
9.83500003814697 2.98873472213745
9.86499977111816 2.98941230773926
9.89500045776367 2.98785352706909
9.92000007629395 2.98465085029602
9.94499969482422 2.97953987121582
9.97000026702881 2.97238063812256
9.99499988555908 2.96309280395508
10.0200004577637 2.95167016983032
10.0450000762939 2.93814396858215
10.0749998092651 2.91925263404846
10.1049995422363 2.89773845672607
10.1400003433228 2.8698627948761
10.1850004196167 2.83081841468811
10.2600002288818 2.76195955276489
10.3299999237061 2.69857048988342
10.3800001144409 2.65620064735413
10.4250001907349 2.62108254432678
10.4650001525879 2.59249591827393
10.5050001144409 2.56637191772461
10.5450000762939 2.54265809059143
10.585000038147 2.52135419845581
10.625 2.5025691986084
10.6599998474121 2.48832941055298
10.6949996948242 2.47620368003845
10.7299995422363 2.46620535850525
10.7650003433228 2.45834183692932
10.8000001907349 2.45263433456421
10.835000038147 2.44911241531372
10.8699998855591 2.44774389266968
10.9049997329712 2.44842648506165
10.9399995803833 2.4510509967804
10.9750003814697 2.45553851127625
11.0150003433228 2.46288919448853
11.0550003051758 2.47259759902954
11.0950002670288 2.48459410667419
11.1400003433228 2.50053954124451
11.1850004196167 2.51876473426819
11.2349996566772 2.54133105278015
11.289999961853 2.56860542297363
11.3500003814697 2.60090327262878
11.4300003051758 2.64677405357361
11.6750001907349 2.78925538063049
11.75 2.82971167564392
11.8149995803833 2.86229825019836
11.875 2.88993668556213
11.9250001907349 2.91076898574829
11.9700002670288 2.92740607261658
12.0150003433228 2.94198822975159
12.0649995803833 2.95594477653503
12.1149997711182 2.96763110160828
12.1700000762939 2.97816181182861
12.2349996566772 2.98817253112793
12.3100004196167 2.99739813804626
12.3950004577637 3.0056746006012
12.4549999237061 3.00953054428101
12.5150003433228 3.01120257377625
12.585000038147 3.0108904838562
12.6549997329712 3.00855326652527
12.7250003814697 3.00398564338684
12.7849998474121 2.99802422523499
12.8400001525879 2.99039459228516
12.8900003433228 2.98127841949463
12.9350004196167 2.97098612785339
12.9799995422363 2.95844841003418
13.0249996185303 2.94354391098022
13.0699996948242 2.92630290985107
13.1199998855591 2.90479612350464
13.1800003051758 2.87651252746582
13.2600002288818 2.83606815338135
13.3549995422363 2.78818249702454
13.4049997329712 2.76518535614014
13.4499998092651 2.7467954158783
13.4899997711182 2.7327139377594
13.5299997329712 2.72095823287964
13.5649995803833 2.71272206306458
13.6000003814697 2.70652627944946
13.6350002288818 2.70244646072388
13.6709995269775 2.70043206214905
13.706000328064 2.70054936408997
13.7410001754761 2.7026948928833
13.7749996185303 2.706702709198
13.8109998703003 2.71297597885132
13.8500003814697 2.72198247909546
13.8900003433228 2.73335075378418
13.9350004196167 2.74830794334412
13.9899997711182 2.76907348632812
14.0600004196167 2.79817533493042
14.1949996948242 2.85735893249512
14.3199996948242 2.91154170036316
14.3950004577637 2.94159340858459
14.4549999237061 2.96340155601501
14.5150003433228 2.98285794258118
14.5799999237061 3.00166249275208
14.6499996185303 3.01977968215942
14.7200002670288 3.03563618659973
14.8050003051758 3.0524365901947
14.8850002288818 3.06624627113342
14.9499998092651 3.07540822029114
15.0200004577637 3.08308887481689
15.1000003814697 3.08974528312683
15.1700000762939 3.09363913536072
15.2349996566772 3.09517669677734
15.3050003051758 3.09476327896118
15.3699998855591 3.09232950210571
15.4409999847412 3.08736085891724
15.5050001144409 3.08082032203674
15.5749998092651 3.07139134407043
15.6599998474121 3.05758571624756
15.9849996566772 3.00185871124268
16.2049999237061 2.96867752075195
16.3250007629395 2.95307159423828
16.4050006866455 2.94471907615662
16.4750003814697 2.93948984146118
16.5349998474121 2.93722939491272
16.5900001525879 2.9375057220459
16.6499996185303 2.94017696380615
16.7299995422363 2.94614505767822
16.9349994659424 2.96442461013794
17.0599994659424 2.97451210021973
17.25 2.98701596260071
17.4950008392334 3.00183582305908
17.5750007629395 3.00413179397583
17.6499996185303 3.00415778160095
17.7159996032715 3.00228762626648
17.7600002288818 2.99915432929993
17.7999992370605 2.99409461021423
17.8349990844727 2.98744559288025
17.871000289917 2.97838997840881
17.9500007629395 2.94782996177673
17.9850006103516 2.93033361434937
18.0249996185303 2.90790939331055
18.0809993743896 2.87373805046082
18.2199993133545 2.78763461112976
18.2700004577637 2.75986862182617
18.3150005340576 2.73729538917542
18.3549995422363 2.71940517425537
18.3959999084473 2.70336079597473
18.4349994659424 2.69027352333069
18.4750003814697 2.67898201942444
18.5149993896484 2.66975259780884
18.5550003051758 2.66256737709045
18.5949993133545 2.65746903419495
18.6350002288818 2.65443992614746
18.6749992370605 2.65343403816223
18.7150001525879 2.6544337272644
18.7549991607666 2.65747356414795
18.7950000762939 2.66257524490356
18.8349990844727 2.66977620124817
18.875 2.67902588844299
18.9200000762939 2.69167351722717
18.9699993133545 2.70808339118958
19.0249996185303 2.72838830947876
19.0949993133545 2.75666475296021
19.2099990844727 2.80582165718079
19.3199996948242 2.85248327255249
19.3899993896484 2.87956190109253
19.4549999237061 2.90243601799011
19.5200004577637 2.92302227020264
19.5849990844727 2.94134187698364
19.6499996185303 2.95741128921509
19.7099990844727 2.97003102302551
19.7749996185303 2.9814760684967
19.8400001525879 2.9908013343811
19.8999996185303 2.99731302261353
19.9699993133545 3.00263929367065
20.0650005340576 3.00742936134338
20.1599998474121 3.01006174087524
20.2800006866455 3.01113080978394
20.4449996948242 3.01233577728271
20.5450000762939 3.0154824256897
20.6299991607666 3.02009892463684
20.8850002288818 3.03641939163208
20.9750003814697 3.03860831260681
21.0400009155273 3.03818130493164
21.1049995422363 3.03553652763367
21.1650009155273 3.03087115287781
21.2299995422363 3.02363038063049
21.3209991455078 3.01105785369873
21.4699993133545 2.99001026153564
21.5450000762939 2.98198676109314
21.6200008392334 2.97626686096191
21.7000007629395 2.97230958938599
21.8150005340576 2.96897721290588
21.9300003051758 2.96771049499512
22.0499992370605 2.96593880653381
22.1149997711182 2.96302843093872
22.1700000762939 2.95858788490295
22.2299995422363 2.95147013664246
22.2999992370605 2.94085359573364
22.3799991607666 2.92635297775269
22.5149993896484 2.89903354644775
22.6459999084473 2.87346291542053
22.7199993133545 2.86118388175964
22.7800006866455 2.85327839851379
22.8299999237061 2.84880638122559
22.875 2.84684228897095
22.9200000762939 2.84697961807251
22.9650001525879 2.84926295280457
23.0100002288818 2.8536422252655
23.0599994659424 2.86063098907471
23.125 2.87210845947266
23.2049999237061 2.88855004310608
23.3600006103516 2.9209361076355
23.4200000762939 2.93102478981018
23.4699993133545 2.93717360496521
23.5149993896484 2.94053983688354
23.5599994659424 2.94171977043152
23.6060009002686 2.94070553779602
23.6499996185303 2.93767189979553
23.7000007629395 2.93190312385559
23.7549991607666 2.92325377464294
23.8299999237061 2.90895462036133
23.9449996948242 2.88709831237793
24.0049991607666 2.87779903411865
24.0550003051758 2.87210655212402
24.1000003814697 2.868976354599
24.1450004577637 2.86788010597229
24.1900005340576 2.86891055107117
24.2350006103516 2.87204051017761
24.2849998474121 2.87768459320068
24.3400001525879 2.88600468635559
24.3999996185303 2.89720177650452
24.4650001525879 2.9114363193512
24.5450000762939 2.93121600151062
24.6749992370605 2.96601819992065
24.7450008392334 2.98389720916748
24.7849998474121 2.99193024635315
24.8199996948242 2.99666547775269
24.8509998321533 2.99855017662048
24.8799991607666 2.99809575080872
24.9099998474121 2.99526309967041
24.9400005340576 2.9901397228241
24.9699993133545 2.98282599449158
25.0049991607666 2.97196173667908
25.0450000762939 2.9570324420929
25.0960006713867 2.93525338172913
25.1849994659424 2.89420294761658
25.2800006866455 2.85117268562317
25.3500003814697 2.82192134857178
25.4099998474121 2.79900240898132
25.4699993133545 2.7784469127655
25.5249996185303 2.76198625564575
25.5750007629395 2.7492778301239
25.6200008392334 2.73988890647888
25.6650009155273 2.7326078414917
25.7099990844727 2.72757291793823
25.7549991607666 2.7248969078064
25.7999992370605 2.72447443008423
25.9449996948242 2.72562313079834
25.9799995422363 2.72276926040649
26.0149993896484 2.71771550178528
26.0450000762939 2.71141362190247
26.0750007629395 2.70309400558472
26.1049995422363 2.69254612922668
26.1350002288818 2.67948341369629
26.1599998474121 2.66641592979431
26.1849994659424 2.65113806724548
26.2099990844727 2.63344311714172
26.2350006103516 2.61313343048096
26.2600002288818 2.59002232551575
26.2849998474121 2.56393694877625
26.3099994659424 2.5347306728363
26.3349990844727 2.50229215621948
26.3600006103516 2.46657109260559
26.3899993896484 2.41954660415649
26.4260005950928 2.35826587677002
26.4549999237061 2.3048734664917
26.4899997711182 2.23519968986511
26.5349998474121 2.13965940475464
26.6049995422363 1.98481428623199
26.7199993133545 1.73028743267059
26.7700004577637 1.6253856420517
26.8099994659424 1.54624140262604
26.8500003814697 1.47253286838531
26.8850002288818 1.41245174407959
26.9150009155273 1.36493098735809
26.9449996948242 1.32176923751831
26.9699993133545 1.28953814506531
26.9950008392334 1.26097118854523
27.0149993896484 1.24088621139526
27.0349998474121 1.22335803508759
27.0550003051758 1.20846915245056
27.0750007629395 1.19629096984863
27.0949993133545 1.18687748908997
27.1149997711182 1.18025922775269
27.1350002288818 1.17643320560455
27.1550006866455 1.1753591299057
27.1749992370605 1.17696714401245
27.1949996948242 1.18116629123688
27.2159996032715 1.18825042247772
27.2350006103516 1.19691455364227
27.2549991607666 1.20823109149933
27.2749996185303 1.22167611122131
27.2999992370605 1.24127721786499
27.3250007629395 1.26375091075897
27.3500003814697 1.28886449337006
27.3799991607666 1.3221949338913
27.4099998474121 1.35870671272278
27.4449996948242 1.40489816665649
27.4850006103516 1.46178722381592
27.5300006866455 1.53014433383942
27.5799999237061 1.61046421527863
27.6399993896484 1.71134209632874
27.8199996948242 2.0177321434021
27.8549995422363 2.0707995891571
27.8859996795654 2.11359572410583
27.9099998474121 2.14340734481812
27.9349994659424 2.17090964317322
27.9549999237061 2.19001817703247
27.9750003814697 2.20634078979492
27.9950008392334 2.21970677375793
28.0149993896484 2.22997307777405
28.0300006866455 2.23556780815125
28.0450000762939 2.23931241035461
28.0610008239746 2.2412269115448
28.0750007629395 2.24111270904541
28.0900001525879 2.23911571502686
28.1049995422363 2.23516607284546
28.1200008392334 2.22925853729248
28.1350002288818 2.22139573097229
28.1499996185303 2.21158766746521
28.1700000762939 2.1955099105835
28.1900005340576 2.17604684829712
28.2099990844727 2.15325903892517
28.2299995422363 2.12722253799438
28.2509994506836 2.09649181365967
28.2749996185303 2.05727958679199
28.2999992370605 2.01201915740967
28.3250007629395 1.96253287792206
28.3549995422363 1.89800357818604
28.3899993896484 1.81667482852936
28.4349994659424 1.70607435703278
28.4750003814697 1.60139906406403
28.5300006866455 1.45016396045685
28.6599998474121 1.09133505821228
28.7150001525879 0.946011543273926
28.7600002288818 0.832988739013672
28.7999992370605 0.737957715988159
28.8400001525879 0.648435354232788
28.875 0.574801921844482
28.9099998474121 0.505749225616455
28.9449996948242 0.441527962684631
28.9750003814697 0.390599012374878
29.0049991607666 0.343750953674316
29.0349998474121 0.301285028457642
29.0599994659424 0.269479870796204
29.0849990844727 0.241143107414246
29.1100006103516 0.216465592384338
29.1299991607666 0.199458837509155
29.1499996185303 0.184922099113464
29.1700000762939 0.172831177711487
29.1900005340576 0.163089752197266
29.2099990844727 0.155521512031555
29.2350006103516 0.148727655410767
29.2600002288818 0.144323468208313
29.2900009155273 0.141272187232971
29.3349990844727 0.13917338848114
29.4109992980957 0.135859847068787
29.4599990844727 0.131221652030945
29.5599994659424 0.120423316955566
29.5849990844727 0.12104058265686
29.6060009002686 0.123937129974365
29.625 0.128758072853088
29.6450004577637 0.136187076568604
29.6650009155273 0.145920634269714
29.6900005340576 0.160907506942749
29.7199993133545 0.182083368301392
29.7600002288818 0.213643789291382
29.8549995422363 0.290304183959961
29.8899993896484 0.31562614440918
29.9249992370605 0.338214635848999
29.9549999237061 0.355085611343384
29.9850006103516 0.369479894638062
30.0149993896484 0.381292104721069
30.0400009155273 0.389125108718872
30.0699996948242 0.396103858947754
30.1000003814697 0.400464415550232
30.1299991607666 0.402254462242126
30.1599998474121 0.401539444923401
30.1900005340576 0.398405194282532
30.2210006713867 0.392741680145264
30.25 0.385339260101318
30.2849998474121 0.373897790908813
30.3199996948242 0.359972834587097
30.3549995422363 0.343863010406494
30.3959999084473 0.32263720035553
30.4449996948242 0.294639587402344
30.5049991607666 0.257726430892944
30.6800003051758 0.148244619369507
30.7310009002686 0.119616150856018
30.7749996185303 0.0971478223800659
30.8199996948242 0.0766712427139282
30.8600006103516 0.0607924461364746
30.8999996185303 0.0471868515014648
30.9400005340576 0.0358470678329468
30.9799995422363 0.0266865491867065
31.0249996185303 0.01878821849823
31.0699996948242 0.013117790222168
31.125 0.00858092308044434
31.1949996948242 0.00525903701782227
31.3400001525879 0.00117480754852295
31.4050006866455 0.00145220756530762
31.4349994659424 0.00369715690612793
31.4659996032715 0.00838935375213623
31.4899997711182 0.0139377117156982
31.5149993896484 0.0216999053955078
31.5400009155273 0.0315961837768555
31.5650005340576 0.0436395406723022
31.5949993133545 0.0607947111129761
31.625 0.0805907249450684
31.6599998474121 0.106395125389099
31.7049999237061 0.142441272735596
31.8099994659424 0.227908134460449
31.8500003814697 0.257187843322754
31.8850002288818 0.280129313468933
31.9200000762939 0.300163507461548
31.9500007629395 0.314832210540771
31.9799995422363 0.327090501785278
32.0099983215332 0.336896181106567
32.0400009155273 0.344253540039062
32.0699996948242 0.349207520484924
32.101001739502 0.351881742477417
32.1300010681152 0.352225303649902
32.1599998474121 0.350491404533386
32.189998626709 0.346744537353516
32.2249984741211 0.339987874031067
32.2599983215332 0.33086371421814
32.2999992370605 0.317927598953247
32.3499984741211 0.299107789993286
32.4500007629395 0.260672926902771
32.4949989318848 0.246181488037109
32.5349998474121 0.235530138015747
32.5800018310547 0.225972771644592
32.6300010681152 0.217756986618042
32.7050018310547 0.208001375198364
32.810001373291 0.194038033485413
32.8800010681152 0.182282686233521
32.9500007629395 0.168289661407471
33.0400009155273 0.1479332447052
33.310001373291 0.085328221321106
33.3600006103516 0.0760993957519531
33.4000015258789 0.0707938671112061
33.435001373291 0.0683026313781738
33.4700012207031 0.0681698322296143
33.5050010681152 0.0705112218856812
33.5400009155273 0.0752189159393311
33.5800018310547 0.0831036567687988
33.625 0.0943541526794434
33.7050018310547 0.1171875
33.7750015258789 0.136220097541809
33.8250007629395 0.147541284561157
33.8699989318848 0.155527830123901
33.9150009155273 0.161213517189026
33.9599990844727 0.164538621902466
34.0050010681152 0.165557622909546
34.0499992370605 0.164402008056641
34.0999984741211 0.160789370536804
34.1500015258789 0.15499222278595
34.2050018310547 0.146438360214233
34.2700004577637 0.133944511413574
34.4000015258789 0.107892990112305
34.435001373291 0.103699684143066
34.4650001525879 0.102296471595764
34.4900016784668 0.103098511695862
34.5149993896484 0.106015682220459
34.5400009155273 0.111315608024597
34.564998626709 0.119216203689575
34.5900001525879 0.129859328269958
34.6150016784668 0.143290042877197
34.6399993896484 0.159446358680725
34.6650009155273 0.17817223072052
34.6949996948242 0.203719139099121
34.7249984741211 0.232219457626343
34.7599983215332 0.26874852180481
34.7949981689453 0.308372735977173
34.8349990844727 0.356986165046692
34.875 0.408894062042236
34.9199981689453 0.471035122871399
34.9700012207031 0.54412317276001
35.060001373291 0.680923581123352
35.1100006103516 0.754977703094482
35.1500015258789 0.810533761978149
35.185001373291 0.855194091796875
35.2150001525879 0.889828443527222
35.2400016784668 0.915734529495239
35.2649993896484 0.938681602478027
35.2849998474121 0.954766035079956
35.310001373291 0.971893906593323
35.3300018310547 0.983142852783203
35.3499984741211 0.992170810699463
35.3699989318848 0.998957395553589
35.3899993896484 1.00349771976471
35.4099998474121 1.00580227375031
35.4300003051758 1.00589561462402
35.4500007629395 1.00381445884705
35.4700012207031 0.999604463577271
35.4900016784668 0.993319272994995
35.5099983215332 0.985020041465759
35.5299987792969 0.974774837493896
35.5550003051758 0.959346413612366
35.5800018310547 0.941170811653137
35.6049995422363 0.920473217964172
35.6349983215332 0.892717361450195
35.6699981689453 0.857110977172852
35.7200012207031 0.802523732185364
35.7949981689453 0.720365762710571
35.8310012817383 0.684184551239014
35.8600006103516 0.657795429229736
35.8849983215332 0.637483835220337
35.9099998474121 0.619741797447205
35.935001373291 0.604805469512939
35.9550018310547 0.594993591308594
35.9749984741211 0.587144136428833
35.9949989318848 0.581279158592224
36.0149993896484 0.577382683753967
36.0349998474121 0.575400114059448
36.0559997558594 0.575277805328369
36.0800018310547 0.577409148216248
36.1049995422363 0.581917285919189
36.1349983215332 0.589897871017456
36.1699981689453 0.601872205734253
36.2200012207031 0.621793031692505
36.2949981689453 0.651878237724304
36.3349990844727 0.665576219558716
36.3699989318848 0.675300717353821
36.4000015258789 0.681622982025146
36.4300003051758 0.68587851524353
36.4599990844727 0.687931776046753
36.4900016784668 0.687695980072021
36.5200004577637 0.685128331184387
36.5499992370605 0.680225849151611
36.5800018310547 0.673005342483521
36.6100006103516 0.663497924804688
36.6399993896484 0.651753425598145
36.6699981689453 0.637836217880249
36.7050018310547 0.618965625762939
36.7400016784668 0.597417116165161
36.7750015258789 0.573390126228333
36.814998626709 0.543190360069275
36.8600006103516 0.506190538406372
36.9150009155273 0.457708954811096
37.0250015258789 0.359435319900513
37.064998626709 0.326783299446106
37.1049995422363 0.296979188919067
37.1349983215332 0.276639938354492
37.1450004577637 0.270241856575012
37.185001373291 0.246461153030396
37.2249984741211 0.225297570228577
37.2700004577637 0.204055309295654
37.3209991455078 0.182481408119202
37.3800010681152 0.159880399703979
37.4500007629395 0.135428547859192
37.5200004577637 0.113237738609314
37.5849990844727 0.0948776006698608
37.6450004577637 0.080125093460083
37.7050018310547 0.0675945281982422
37.7649993896484 0.0572090148925781
37.8300018310547 0.0481135845184326
37.9049987792969 0.0398664474487305
37.9949989318848 0.0321907997131348
38.125 0.0234392881393433
38.2999992370605 0.0137554407119751
38.3549995422363 0.0124303102493286
38.4000015258789 0.013278603553772
38.4360008239746 0.015907883644104
38.4860000610352 0.0238209962844849
38.5209999084473 0.0320993661880493
38.560001373291 0.0437709093093872
38.6049995422363 0.0598030090332031
38.6699981689453 0.0857740640640259
38.814998626709 0.146525621414185
38.875 0.174334049224854
38.935001373291 0.204780220985413
39.0250015258789 0.253591060638428
39.0900001525879 0.287964582443237
39.1349983215332 0.30955171585083
39.1749992370605 0.326390504837036
39.2099990844727 0.338881731033325
39.2449989318848 0.348986864089966
39.2750015258789 0.355587244033813
39.3050003051758 0.360191345214844
39.3349990844727 0.36275851726532
39.3650016784668 0.363285899162292
39.3950004577637 0.361804366111755
39.4249992370605 0.358377933502197
39.4599990844727 0.35204291343689
39.4949989318848 0.343356013298035
39.5299987792969 0.332514047622681
39.5699996948242 0.317762136459351
39.6150016784668 0.298575162887573
39.6650009155273 0.274665594100952
39.7249984741211 0.243413686752319
39.8559989929199 0.172224044799805
39.8650016784668 0.166827201843262
39.935001373291 0.130393505096436
39.9900016784668 0.104084849357605
40.0400009155273 0.0825127363204956
40.0849990844727 0.065316915512085
40.1300010681152 0.0503965616226196
40.1749992370605 0.0378297567367554
40.2200012207031 0.0275992155075073
40.265998840332 0.0194423198699951
40.314998626709 0.0130679607391357
40.3650016784668 0.00864636898040771
40.4300003051758 0.00524449348449707
40.5299987792969 0.00267946720123291
40.6650009155273 0.000993132591247559
40.8450012207031 0.00543880462646484
40.939998626709 0.00505721569061279
41.0299987792969 0.002693772315979
41.0849990844727 0.00109124183654785
41.125 0.00606989860534668
41.1599998474121 0.0124766826629639
41.1949996948242 0.0209485292434692
41.2350006103516 0.0330754518508911
41.2799987792969 0.0493602752685547
41.3349990844727 0.0717778205871582
41.4500007629395 0.119211435317993
41.5 0.136953473091125
41.5449981689453 0.150417923927307
41.5849990844727 0.160143256187439
41.6399993896484 0.170761704444885
41.689998626709 0.178575992584229
41.7649993896484 0.18994128704071
41.7999992370605 0.197407960891724
41.8300018310547 0.205834031105042
41.8549995422363 0.214817523956299
41.8800010681152 0.226033091545105
41.9049987792969 0.239810705184937
41.9300003051758 0.25631582736969
41.9550018310547 0.275524377822876
41.9809989929199 0.29816746711731
42.0099983215332 0.326209425926208
42.0449981689453 0.36310350894928
42.0999984741211 0.424951791763306
42.1699981689453 0.503466010093689
42.2099990844727 0.544939279556274
42.2449989318848 0.577823758125305
42.2750015258789 0.602922677993774
42.2999992370605 0.62139880657196
42.326000213623 0.63807487487793
42.3499984741211 0.651042342185974
42.375 0.661972522735596
42.4000015258789 0.670206546783447
42.4249992370605 0.675719022750854
42.4500007629395 0.678523540496826
42.4749984741211 0.67866837978363
42.4949989318848 0.676918268203735
42.5200004577637 0.672476053237915
42.5449981689453 0.66562819480896
42.5709991455078 0.656080007553101
42.5950012207031 0.645193219184875
42.625 0.62897777557373
42.6549987792969 0.610101938247681
42.685001373291 0.588829278945923
42.7200012207031 0.561346769332886
42.7599983215332 0.526969313621521
42.8050003051758 0.485313653945923
42.8650016784668 0.426485180854797
43 0.291217684745789
};
\end{axis}

\end{tikzpicture}

%% file: figures/cave-sim-table.tex
\begin{tabular}{lcccccccc}
  \toprule
  \multirow{3}{*}{Method} &
    \multicolumn{2}{c}{Region 1} &
    \multicolumn{2}{c}{Region 2} &
    \multicolumn{2}{c}{Region 3} &
    \multicolumn{2}{c}{Return} \\
    & {Time} & {Av. Speed} & {Time} & {Av. Speed} & {Time} & {Av. Speed} & {Time} & {Av. Speed} \\
    & {(s)} & {(m/s)} & {(s)} & {(m/s)} & {(s)} & {(m/s)} & {(s)} & {(m/s)} \\
    \midrule
  No adapt. \SI{0.5}{\metre} & \textbf{20.0} & 2.22 & -- & -- & -- & -- & -- & --\\
  No adapt. \SI{0.2}{\metre} & 42.4 & 1.15 & 43.2 & 0.98 & 26.7 & \textbf{0.97} & 25.3 & 1.13\\
  \textbf{Adaptation} & \textbf{20.0} & \textbf{2.67} & \textbf{43.0} & \textbf{1.02} & \textbf{22.3} & 0.93 & \textbf{17.7} & \textbf{1.69}\\
  \bottomrule
\end{tabular}

%% file: figures/hw_speed.tex
\begin{tikzpicture}

\definecolor{darkgray176}{RGB}{176,176,176}
\definecolor{darkgreen}{RGB}{0,100,0}
\definecolor{lightgray204}{RGB}{204,204,204}
\definecolor{maroon}{RGB}{128,0,0}
\definecolor{navy}{RGB}{0,0,128}

\begin{axis}[
width=\textwidth,
height=5cm,
legend cell align={left},
legend style={
  nodes={scale=0.5, transform shape},
  cells={align=left},
  fill opacity=0.8,
  draw opacity=1,
  text opacity=1,
  at={(0.99,0.01)},
  anchor=south east,
  draw=lightgray204
},
tick align=outside,
tick pos=left,
x grid style={darkgray176},
xlabel={X (m)},
xmin=-3.5, xmax=3.0,
xtick style={color=black},
y grid style={darkgray176},
ylabel={Speed (m/s)},
ymin=0, ymax=3.1,
ytick style={color=black}
]
\addplot [thick, maroon]
table {%
-3.50186160178299 1.72119265457111
-3.48466185761911 1.71874667955123
-3.4674868579465 1.71624377946492
-3.45033717125676 1.71368411257912
-3.43321336426565 1.71106787575072
-3.41611600152848 1.70839530417108
-3.39904564505811 1.70566667108677
-3.38200285394592 1.70288228749716
-3.3649881839853 1.70004250211869
-3.3480021872837 1.69714770413289
-3.3310454118426 1.69419832556257
-3.3141184011318 1.69119484055822
-3.29722169367312 1.68813776438491
-3.28035582263445 1.68502765241018
-3.26352131543363 1.68186509909325
-3.2467186933527 1.67865073697553
-3.22994847116207 1.67538523567276
-3.21321115426324 1.67206930037098
-3.19650724830742 1.66870367131558
-3.17983724396531 1.66528910370624
-3.16320162684157 1.66182635360797
-3.14660087498201 1.65831617687587
-3.13003545887119 1.65475933005172
-3.11350584142126 1.65115657121876
-3.0970124779524 1.64750866081495
-3.08055581616541 1.64381636240563
-3.06413629610673 1.64008044341578
-3.04775434768807 1.63630167525652
-3.03141040039641 1.63248083623303
-3.01510486859247 1.62861870869643
-2.99883816122154 1.62471608165642
-2.98261067930701 1.62077375053865
-2.96642281588469 1.61679251747748
-2.95027495593437 1.61277319158419
-2.93416747630883 1.60871658919112
-2.91810074566048 1.60462353407197
-2.90207512436597 1.60049485763876
-2.88609096207017 1.59633139849275
-2.87014860718643 1.59213397612538
-2.85424839407162 1.5879030748605
-2.83839065767414 1.58363855783297
-2.82257573738972 1.57933967947785
-2.80680398293933 1.57500512714904
-2.79107575983673 1.57063306158464
-2.77539145445705 1.56622115623154
-2.75975147871779 1.56176663544197
-2.74415627438344 1.55726631155479
-2.72860631700479 1.55271662087407
-2.71310211943791 1.54811369150279
-2.69764423332597 1.54345373257812
-2.68223324610237 1.53873339402635
-2.66686977640798 1.53394978127102
-2.65155446963015 1.52910043650247
-2.63628799362709 1.52418332033372
-2.62107103463376 1.51919679383901
-2.60590429334551 1.5141396009717
-2.59078848117557 1.50901085135801
-2.57572431668288 1.50381000346323
-2.56071252217465 1.49853684393801
-2.54575382068732 1.49319142628637
-2.53084893365492 1.48777401336269
-2.51599857889596 1.48228506390066
-2.50120346869973 1.4767252235268
-2.48646430800156 1.4710953159605
-2.47178179264507 1.46539633439878
-2.45715660772974 1.45962943308434
-2.44258942604184 1.45379591905508
-2.42808090440947 1.4478972431898
-2.41363169094767 1.4419349889816
-2.39924241112381 1.43591078541256
-2.3849136763945 1.42982623422563
-2.3635353949825 1.4205896794085
-2.35644020910383 1.41748233444194
-2.3422966214392 1.41122603461802
-2.32821586922783 1.40491548908326
-2.31419848760707 1.39855215795376
-2.30024499722495 1.39213747929543
-2.28635590445112 1.38567287103972
-2.27253170158196 1.37915972628147
-2.25877286736471 1.37259934426859
-2.24507986833276 1.36599286626731
-2.2314531603864 1.35934127935805
-2.21789319028339 1.35264542664276
-2.20440039702892 1.34590601716915
-2.19097521316763 1.33912363557486
-2.17761806598074 1.33229875145479
-2.16432937859068 1.32543172845443
-2.1511095690116 1.31852283205984
-2.13795905894929 1.31157224108528
-2.1248782628898 1.30458003680043
-2.1118675969695 1.29754620260324
-2.09892747795489 1.29047063049673
-2.0860583241141 1.28335312954072
-2.07326055600512 1.27619343404769
-2.06053459718334 1.26899121152565
-2.04788087483097 1.26174607037095
-2.03529982031066 1.25445756731384
-2.02279186964599 1.24712521461972
-2.01035746392727 1.23974848895872
-1.99799704955102 1.23232685865989
-1.98571107815499 1.22485980880716
-1.97350000642065 1.21734684603079
-1.96136429584418 1.20978750120758
-1.9493044124809 1.20218133204105
-1.93732082666471 1.19452792552316
-1.92541401270329 1.18682690027915
-1.91358444855059 1.17907790879698
-1.90183261371213 1.17128063837747
-1.89015899586987 1.16343481698415
-1.87856408002912 1.1555401993489
-1.86704835527499 1.14759656491783
-1.85561231277486 1.13960371816405
-1.8442564455171 1.13156149077034
-1.8329812480285 1.12346974370638
-1.82178721607146 1.11532836920179
-1.81067484632194 1.10713729261647
-1.79964463602926 1.09889647420966
-1.78869708265864 1.09060591080908
-1.77783268351498 1.08226563867472
-1.76705193528515 1.07387574887635
-1.75635533340447 1.06543640127628
-1.74574337136334 1.05694782482835
-1.73521654002154 1.0484103165234
-1.7247753269332 1.03982424031554
-1.71442021568268 1.0311900260295
-1.70415168523133 1.02250816824986
-1.69397020927556 1.01377922519238
-1.68387625411864 1.00500381624711
-1.67387028405346 0.996182626894858
-1.66395275168545 0.987316408727287
-1.65412410321688 0.978405991675604
-1.64438477617256 0.969452280855785
-1.63473519866138 0.96045625388285
-1.62517578866468 0.951418958217924
-1.61570695335088 0.942341508547969
-1.60632908841622 0.933225084198144
-1.59704257745111 0.924070926576676
-1.58784779133203 0.914880336652177
-1.57874508763736 0.905654672953471
-1.5697348100629 0.896395354380082
-1.5608172878006 0.887103862552221
-1.55199283492376 0.877781739043247
-1.54326174980358 0.868430582166432
-1.53462431455761 0.859052043821083
-1.52608079452936 0.849647826397619
-1.51763143779871 0.840219679741158
-1.50927647472219 0.830769398173215
-1.50101611627916 0.821298816158414
-1.49285055857949 0.811809811317945
-1.48477997509326 0.802304295613968
-1.47680452120259 0.792784220470108
-1.46892433259688 0.783251572492585
-1.46113952493625 0.773708370433604
-1.45345019354497 0.764156662222713
-1.44585641313422 0.754598522065537
-1.43835823755361 0.745036047609306
-1.43095569957062 0.735471357174594
-1.42364881067745 0.725906587052683
-1.41643756092444 0.716343888912733
-1.40932191877732 0.706785427709994
-1.40230183099419 0.697233379612353
-1.39537722252575 0.687689929519284
-1.38854799644032 0.678157268599917
-1.38181403387321 0.668637591895783
-1.37517519399967 0.659133095987635
-1.36863131403097 0.649645976725706
-1.36218220923273 0.640178427022808
-1.35582767296509 0.630732634709644
-1.34956747674415 0.621310780376588
-1.34340137032779 0.61191503454493
-1.33732908183076 0.602547554973338
-1.33135031786181 0.593210484811985
-1.32546476367839 0.583905950889103
-1.31967208335823 0.574636062051902
-1.31397191998719 0.565402907561321
-1.30836389586293 0.556208555540026
-1.30284761271386 0.547055051473087
-1.29742265113118 0.537944415406485
-1.29208857403695 0.52887864583124
-1.2868449221031 0.519859708646739
-1.28169121729895 0.510889539901605
-1.27662696239424 0.501970043296805
-1.27165164128739 0.493103089239536
-1.26676471934303 0.484290513937856
-1.26196564373831 0.475534118535594
-1.25725384381773 0.466835668287062
-1.25262873145597 0.458196891771111
-1.24808970075846 0.449619478870592
-1.24363613113212 0.44110508499261
-1.23926738362131 0.432655320859389
-1.23498280404816 0.424271755302085
-1.23078172276449 0.415955913513003
-1.22666345508945 0.407709276787249
-1.22262730174972 0.399533282290788
-1.21867254932183 0.391429322854557
-1.21479847067642 0.383398746794286
-1.21100432542422 0.375442857755681
-1.20728936036342 0.367562914584623
-1.20365280992853 0.359760131045438
-1.20009389664927 0.352035673845507
-1.19661183163215 0.344390660860623
-1.19320581504881 0.336826161235165
-1.18987503662187 0.329343195677729
-1.18661867610778 0.321942736769882
-1.18343590377645 0.314625709287813
-1.18032588088768 0.307392990536654
-1.17728776016415 0.300245410697258
-1.17432068626082 0.293183753185197
-1.17142379623116 0.286208754861135
-1.16859621999761 0.279321104691484
-1.1658370808384 0.272521442581355
-1.1631454958759 0.265810359901475
-1.16052057655859 0.259188400182791
-1.15796142913593 0.252656059813024
-1.15546715512628 0.246213788735061
-1.15303685177792 0.239861991147062
-1.15066961252293 0.233601026204178
-1.14836452742412 0.227431208721763
-1.14612068361519 0.221352809746061
-1.14393716574066 0.215366055825477
-1.14181305640539 0.20947112841698
-1.13974743662188 0.203668164695161
-1.13773938624838 0.197957258492775
-1.1357879844176 0.192338461234645
-1.13389230995621 0.186811782864913
};
\addlegendentry{no adapt.\\0.6m}
\addplot [thick, navy]
table {%
-3.4994618312974 1.46510685998903
-3.48479671677075 1.46791813304118
-3.47010342547186 1.47074235367009
-3.45538182383952 1.47358032983463
-3.44063177050551 1.47643281510854
-3.42585311683462 1.47930050944088
-3.41104570745707 1.48218405992509
-3.38878032141457 1.4865404034849
-3.38134396957029 1.48800105811713
-3.35899094064533 1.49240948327429
-3.34405205344101 1.49537101070858
-3.32908345880162 1.49835102514958
-3.3140849703032 1.50134982823098
-3.29905639877745 1.5043676674211
-3.28399755284611 1.50740473750154
-3.2689082394404 1.5104611820307
-3.2537882643064 1.51353709460359
-3.23863743250571 1.51663251825627
-3.22345554892533 1.51974744504332
-3.20824241878037 1.52288181758223
-3.19299784809975 1.52603553076695
-3.17772164419508 1.52920843345619
-3.16241361611258 1.53240033013637
-3.14707357506864 1.53561098255974
-3.13170133486908 1.53884011135736
-3.11629671231267 1.54208739747726
-3.10085952758666 1.54535248213391
-3.08538960466546 1.54863496688363
-3.06988677169955 1.55193441525327
-3.05435086138699 1.55525035449095
-3.03878171132756 1.55858227728449
-3.02317916435994 1.5619296434476
-3.00754306888211 1.56529188157413
-2.99187327915548 1.56866839066067
-2.97616965559279 1.57205854169763
-2.96043206503058 1.57546167915755
-2.94466038098974 1.5788771217552
-2.92885448392989 1.58230416324895
-2.9130142614916 1.58574207396334
-2.89713960872279 1.58919010235112
-2.88123042828987 1.59264747652129
-2.86528663067362 1.59611340573343
-2.84930813435039 1.59958708185856
-2.83329486595876 1.6030676808069
-2.81724675805827 1.60655436444275
-2.8011637588388 1.61004627986737
-2.78504581789784 1.61354256386391
-2.76889289591297 1.61704234211596
-2.75270496235254 1.62054473099366
-2.7364819955626 1.62404883878965
-2.72022398284165 1.62755376692387
-2.70393092050342 1.63105861111757
-2.68760281392817 1.63456246253672
-2.67123967760259 1.63806440890523
-2.65484153270275 1.64156353610937
-2.63840841689147 1.64505892721154
-2.62194036966278 1.6485496670917
-2.6054374421005 1.65203484179931
-2.58889969441552 1.65551354000719
-2.57232719591532 1.65898485387103
-2.55572002496491 1.66244787986304
-2.53907826893962 1.66590171958037
-2.52240202416994 1.66934548052828
-2.50569139587854 1.67277827687847
-2.48894649811003 1.67619923020279
-2.47216745365314 1.67960747025339
-2.4553543939525 1.68300213640889
-2.43850745900471 1.68638237902577
-2.42162679724539 1.68974736002277
-2.40471256543114 1.69309625337283
-2.38776492851654 1.69642824557607
-2.37078405952666 1.69974253611369
-2.35377013942503 1.70303833788338
-2.33672335697736 1.70631487761613
-2.31964390606378 1.70957139675856
-2.30253199571941 1.71280714999636
-2.28538783471189 1.71602140939616
-2.26821164155076 1.71921346359318
-2.25100364177279 1.72238261854468
-2.23376406776781 1.72552819770074
-2.21649315860306 1.72864954216241
-2.19919115984588 1.73174601082768
-2.18185832338507 1.73481698052524
-2.16449490725101 1.73786184613619
-2.14710117543452 1.74088002070399
-2.12967739770461 1.74387093561245
-2.11222384942116 1.74683404152855
-2.09474081133691 1.74976880925402
-2.07722856939581 1.7526747297292
-2.05968741453206 1.755551313948
-2.04211764247006 1.75839809286646
-2.0245195535252 1.76121461730498
-2.00689345240581 1.76400045784427
-1.98923964801605 1.76675520471513
-1.97155845062328 1.76947846808566
-1.95385018220609 1.77216987639231
-1.93611516044869 1.77482907941375
-1.91835370907086 1.77745574720465
-1.90056615498913 1.7800495703106
-1.88275282811325 1.78261025949941
-1.86491406114532 1.78513754549098
-1.84705018938158 1.78763117868581
-1.82916155051702 1.79009092889214
-1.81124848445265 1.79251658505184
-1.79331133043092 1.79490795531882
-1.77535043754335 1.79726486542161
-1.75736614850586 1.7995871611578
-1.73935881015442 1.80187470725968
-1.7213287705825 1.80412738742887
-1.70327637895334 1.80634510395332
-1.68520198531592 1.80852777732659
-1.66710594042485 1.81067534586961
-1.64898859556395 1.81278776535485
-1.63085030237363 1.81486500863288
-1.61269141268194 1.81690706526138
-1.59451227833927 1.8189139411817
-1.57631325105438 1.82088565883586
-1.55809468222851 1.82282225724079
-1.53985692279148 1.82472379159325
-1.52160032304206 1.82659033283319
-1.50332523249267 1.82842196721224
-1.4850319997185 1.83021879586709
-1.46672097221064 1.83198093439795
-1.44839249623364 1.83370851245195
-1.43004691395209 1.8354016735612
-1.41168457423362 1.83706057376334
-1.39330581611468 1.8386853830459
-1.37491097960109 1.84027628423649
-1.3565004028066 1.84183347284236
-1.33807442182835 1.84335715660954
-1.31963337062663 1.84484755508842
-1.301177580909 1.84630489920604
-1.28270738201861 1.84772943084471
-1.26422310082675 1.84912140242717
-1.24572506162949 1.85048107650815
-1.22721358604831 1.85180872539068
-1.20868899293387 1.85310463092623
-1.19015159827131 1.85436908430282
-1.17160171508886 1.85560238564809
-1.15303965337049 1.85680484362057
-1.1344657199726 1.85797677500829
-1.11588021854468 1.85911850433439
-1.09728344945398 1.8602303634699
-1.07867570971385 1.86131269125355
-1.06005729014086 1.86236583327329
-1.04142848638965 1.8633901408845
-1.02278958224588 1.86438597187235
-1.00414086066386 1.86535368954007
-0.985482600937933 1.86629366251676
-0.966815078650997 1.86720626440521
-0.948138565626495 1.86809187343687
-0.929453329883819 1.86895087213387
-0.910759635597067 1.8697836469779
-0.89205774305708 1.87059058808615
-0.873347905848116 1.87137208900861
-0.854630381969336 1.87212854595492
-0.835905417078354 1.87286035817915
-0.81717325561665 1.87356792720395
-0.798434137999743 1.87425165664944
-0.779688300597646 1.8749119519509
-0.760935975718096 1.87554922008294
-0.742177391592515 1.87616386929017
-0.72341277236462 1.87675630882411
-0.70464233808166 1.87732694868644
-0.685866304688155 1.87787619937851
-0.667084884022151 1.87840447165009
-0.648298283814231 1.87891217618592
-0.629506707689755 1.87939972330526
-0.610710355173646 1.8798675227411
-0.591909421697306 1.8803159834321
-0.573104098607551 1.88074551332003
-0.554294573177604 1.88115651915257
-0.535481028619971 1.88154940629166
-0.516663644101252 1.8819245785271
-0.497842591953949 1.88228243794752
-0.479018048913668 1.88262338454315
-0.460190179312477 1.88294781623787
-0.441359146320387 1.88325612840163
-0.422525109166841 1.88354871375309
-0.403688223167938 1.88382596222304
-0.384848639754944 1.88408826082231
-0.366006506504164 1.88433599351393
-0.347161967168018 1.88456954108962
-0.328315161707375 1.88478928105054
-0.309466226325016 1.88499558749217
-0.29061529350029 1.88518883098063
-0.271762492025497 1.88536937830722
-0.252907947044935 1.88553759226002
-0.234051780095426 1.88569383153497
-0.215194109147587 1.88583845066348
-0.196335048647841 1.8859717999432
-0.177474709561075 1.88609422537203
-0.15861319941396 1.88620606858525
-0.139750622338884 1.88630766679569
-0.120887076307486 1.88639935275003
-0.102022664419474 1.88648145460866
-0.083157478083546 1.8865542958331
-0.0642916083092793 1.88661819495183
-0.0454251429457226 1.88667346554006
-0.0265581667310162 1.88672041620113
-0.00769076134218594 1.8867593505498
0.0111769945548676 1.88679056719778
0.0300450252553102 1.88681435974125
0.0489132579651397 1.88683101675043
0.0677816227508856 1.88684082176113
0.0866500524892189 1.88684405325566
0.105518482815828 1.8868409845216
0.1243868520727 1.88683188352533
0.143255101254898 1.88681701292099
0.162123173957562 1.88679663007373
0.180991016323148 1.88677098708406
0.199858576988886 1.88674033081332
0.218725807034549 1.88670490291022
0.23759265993043 1.88666493983854
0.256459091485684 1.88662067290584
0.275325059796828 1.88657232828191
0.294190525196079 1.88652012690843
0.313055450198535 1.88646428442081
0.331919799449267 1.88640501118957
0.35078353967095 1.88634251237367
0.369646639612027 1.88627698797412
0.388509069995374 1.88620863288799
0.407370803467543 1.8861376369626
0.426231814548551 1.88606418505014
0.445092079582171 1.8859884570625
0.463951576686823 1.88591062801702
0.482810285706464 1.88583086799101
0.501668188160922 1.885749342086
0.520525267196421 1.88566621049072
0.539381507536794 1.88558162855353
0.558236895435474 1.88549574685454
0.577091418628141 1.88540871127735
0.595945066286131 1.88532066308064
0.614797828970536 1.88523173896928
0.633649698587027 1.88514207116527
0.652500668341363 1.88505178747095
0.671350732695222 1.88496101126083
0.69019988732183 1.88486986148031
0.709048129062035 1.88477845272082
0.72789545588118 1.8846868953018
0.74674186682676 1.88459529535187
0.765587361986956 1.88450375488928
0.784431942449899 1.88441237190163
0.803275610263787 1.88432124042485
0.822118371205454 1.88423045060797
0.84096022351101 1.88414008883939
0.859801174684709 1.88405023768622
0.878641230229942 1.88396097597341
0.897480396419873 1.88387237884948
0.916318680261138 1.88378451787025
0.935156089458296 1.88369746108147
0.953992632379164 1.88361127310049
0.972828318020927 1.88352601519692
0.991663155977053 1.88344174537223
1.01049715921144 1.88335851842612
1.02933033280109 1.88327638608024
1.04816269074408 1.88319539694443
1.06699424470298 1.88311559660961
1.08582500678139 1.88303702771249
1.10465498949487 1.88295973001514
1.12348420574254 1.88288374048337
1.14231266877954 1.88280909336396
1.16114039219016 1.88273582026076
1.17996738986187 1.88266395020963
1.19879367595994 1.88259350975234
1.21761926490289 1.88252452300749
1.23644417133855 1.88245701172371
1.25526841012059 1.88239099533336
1.27409199628572 1.88232649102333
1.2929149450316 1.88226351380629
1.31173727169542 1.88220207659081
1.33055899173322 1.8821421902504
1.34938012069991 1.88208386369122
1.36820067422987 1.8820271039188
1.38702067082269 1.88197191609539
1.42465904215242 1.8818662682211
1.42465904215242 1.8818662682211
1.44347745122911 1.88181580988172
1.46229536360194 1.88176692705868
1.48111279501186 1.88171961666012
1.4999297611615 1.88167387412162
1.51874627770072 1.8816296934646
1.53756236021276 1.88158706735362
1.55637802420097 1.88154598715263
1.57519328507604 1.88150644298012
1.59400815814388 1.88146842376371
1.61282265859394 1.88143191729961
1.63163680148824 1.88139691031027
1.65045060175084 1.88136338849488
1.6692640741579 1.88133133657826
1.68807723332816 1.88130073835867
1.70689009371388 1.88127157675477
1.72570266959234 1.88124383385146
1.74451497505765 1.8812174909449
1.76332702681634 1.88119252858282
1.78213883296728 1.88116892662352
1.80095040981423 1.88114666426934
1.81976177064679 1.88112572013534
1.83857292853793 1.88110607228477
1.85738389633865 1.88108769826612
1.87619468667298 1.88107057514946
1.89500531193344 1.88105467956186
1.91381578427675 1.88103998772199
1.93262611561989 1.88102647547401
1.95143631763658 1.88101411832049
1.97024640175397 1.88100289145636
1.98905637914982 1.88099276981835
2.00786626075019 1.88098372813212
2.02667605722746 1.88097574094073
2.04548577899863 1.88096878263091
2.06429543622388 1.88096282745849
2.08310503880543 1.88095784957329
2.10191459638654 1.88095382304328
2.12072411835085 1.88095072187814
2.13953361662473 1.88094852005188
2.15834309446573 1.88094719152816
2.17715256327961 1.88094671029993
2.19596203140958 1.88094705043011
2.2147715069399 1.88094818607014
2.23358099769685 1.88095009147653
2.25239051124989 1.88095274102694
2.2712000549129 1.88095610923573
2.29000963574567 1.88096017076906
2.30881926055545 1.88096490045949
2.3276289387016 1.88097027332091
2.34643867078594 1.88097621465352
2.36524846280503 1.88098206118332
2.38405830597983 1.88098606332642
2.40286816865565 1.88098540763359
2.42167798602332 1.88097628844222
2.44048765054719 1.88095397757542
2.45929700308046 1.88091289210961
2.47810582464853 1.88084666023161
2.49691382888156 1.8807481852067
2.51572065507725 1.88060970747874
2.53452586196999 1.88042291207332
2.55332892451691 1.88017949559263
2.57212923718969 1.87987169031104
2.59092611962546 1.87949229396649
2.60971882218965 1.87903465144931
2.62850653135722 1.87849263681325
2.64728837491578 1.87786063560615
2.6660634269934 1.87713352751771
2.68483071291469 1.87630666934181
2.70358921388786 1.87537587825089
2.72233787147815 1.87433739138547
2.7410755906837 1.87318759041607
2.75980123981893 1.87192274525027
2.77851364929189 1.87053900769539
2.79721161051208 1.86903242913834
2.81589387497236 1.86739897767042
2.83455915349932 1.86563455466325
2.85320611566669 1.86373501080223
2.87183338936641 1.86169616158378
2.89043956330295 1.85951380194635
2.90902317577089 1.85718371917521
2.9275827311492 1.85470167372472
2.94611668843486 1.85206338412704
2.96462346364882 1.84926453916194
2.98310142961324 1.84630081272512
};
\addlegendentry{no adapt.\\0.3m}
\addplot [thick, darkgreen]
table {%
-3.50752835717453 2.11947728888408
-3.48629472595565 2.12726670411434
-3.46498267548902 2.13516039905408
-3.44359118521928 2.14315391949924
-3.42211927929994 2.15124277876128
-3.40056602689216 2.15942246284874
-3.37893054241245 2.16768843551523
-3.35721198573049 2.17603614317511
-3.33540956231856 2.18446101968805
-3.31352252008537 2.19295849253995
-3.29155016247705 2.20152398504039
-3.26949182887484 2.21015293081139
-3.24734690754164 2.21884077281218
-3.22511483224228 2.22758296830689
-3.20279508208235 2.23637499246216
-3.18038718131172 2.24521234183675
-3.15789069909356 2.25409053776425
-3.13530524924011 2.26300512963002
-3.11263048991628 2.27195169804336
-3.08986612006582 2.28092578301862
-3.06701189670907 2.28992205849299
-3.04406762789024 2.29893358294871
-3.02103320247446 2.30795183109433
-2.99790860202363 2.31696680729039
-2.9746939149523 2.32596715455968
-2.95138934960869 2.33494026055096
-2.92799524631122 2.3438723604896
-2.90451208837038 2.35274863714874
-2.88094051212569 2.36155331787449
-2.85728131602703 2.37026976869863
-2.83353546878918 2.37888058557245
-2.80970411664835 2.38736768275512
-2.7857885897489 2.39571237839027
-2.76179040768818 2.40389547730418
-2.73771128424717 2.4118973510592
-2.71355313133418 2.41969801529582
-2.68931806216849 2.42727720439714
-2.6650083937307 2.4346144435089
-2.64062664850581 2.44168911794902
-2.61617555535326 2.44848063582944
-2.5916580447984 2.454969525227
-2.56707723165446 2.46113844410092
-2.54243639339841 2.46697218655554
-2.51773894926212 2.47245759262658
-2.4929884402161 2.47758345999842
-2.46818850982644 2.48234045763469
-2.44334288596596 2.48672104130414
-2.41845536336069 2.49071937098421
-2.39352978323616 2.49433123063363
-2.36857003235345 2.49755394114082
-2.3435800059875 2.50038622124925
-2.31856360956243 2.50282805092421
-2.29352474268661 2.50488060303698
-2.26846728753073 2.50654618216264
-2.24339509781094 2.50782816479365
-2.21831198836269 2.50873094095571
-2.19322172529159 2.50925985721224
-2.16812801668725 2.50942116104381
-2.14303450388651 2.50922194658905
-2.1179447496875 2.50867002522067
-2.09286224970841 2.50777304931453
-2.06779041650489 2.50653769943778
-2.04273260454879 2.50496967002166
-2.01769211603941 2.5030737313311
-1.99267220984225 2.50085378978104
-1.9676761098329 2.49831294656208
-1.9427070126632 2.49545355459435
-1.917768094966 2.49227727382851
-1.89286252001523 2.48878512491267
-1.86799344370092 2.48497761926814
-1.8431640159791 2.48085565183675
-1.81837737199057 2.47642132716678
-1.79363661967108 2.47167797032873
-1.76894482788501 2.46663005941612
-1.74430501522587 2.46128315966051
-1.71972013946767 2.45564385914487
-1.69519308765127 2.44971970609946
-1.67072666678989 2.44351914776447
-1.6463235951782 2.43705147080346
-1.62198649082142 2.43032666576655
-1.59771788248809 2.42335454892453
-1.57352019470865 2.41614393856976
-1.54939577257177 2.40870263981774
-1.52534688794476 2.40103750132238
-1.50137574875601 2.39315447143897
-1.47748450772356 2.385058652868
-1.45367527054468 2.3767543557969
-1.42995010356147 2.36824514955608
-1.40631104091728 2.3595339128061
-1.38276009121378 2.35062288706038
-1.35929924256421 2.34151408292456
-1.33593045936109 2.33221037616091
-1.31265567135776 2.32271581119739
-1.2894767622811 2.31303553765781
-1.26639555911411 2.30317574371389
-1.24341382203686 2.29314359109192
-1.22053323500903 2.28294715171723
-1.19775539697796 2.27259534597998
-1.17508181369596 2.26209788260561
-1.15251388678085 2.25146519370637
-1.13005292127449 2.24070802327602
-1.10770010801281 2.22983630613311
-1.08545654746574 2.21885885931378
-1.06332325730747 2.20778343463091
-1.04130118277996 2.19661677617723
-1.01939120648887 2.18536467633925
-0.997594157646622 2.17403203033823
-0.975910820777487 2.16262288931388
-0.954341943898971 2.15114051196683
-0.932888246044548 2.13958748931207
-0.911550420457494 2.12796659833008
-0.890329127902979 2.116281592812
-0.869224986608768 2.10453721593718
-0.848238562684024 2.09273913799993
-0.827370361152949 2.08089389571247
-0.806620817587578 2.06900883306782
-0.785990290324229 2.05709204374683
-0.765479053248204 2.0451523150533
-0.745087289131583 2.03319907336112
-0.724815080650779 2.02124225263116
-0.704662420946732 2.00929142234834
-0.684629204169956 1.99735496227561
-0.664715249731157 1.98544005536826
-0.644920310942129 1.97355274878574
-0.625244086066358 1.96169801510292
-0.605686228765324 1.94987981195562
-0.586246357956139 1.9381011401371
-0.566924067095876 1.9263641001622
-0.547718932907856 1.91466994731549
-0.528630523564968 1.90301914520031
-0.509658406344911 1.89141141780518
-0.490802154772139 1.87984580010412
-0.472061355261039 1.86832068720766
-0.453435613274788 1.85683388208103
-0.434924559014137 1.84538264184611
-0.416527852650151 1.83396372268389
-0.398245189114896 1.82257342335394
-0.380076302463738 1.81120762734743
-0.362020969822911 1.79986184369056
-0.344079014778569 1.78853132480687
-0.32625030634659 1.77721195080828
-0.308534747690383 1.76590104649664
-0.290932261444157 1.75459737933314
-0.273442775690368 1.74330107895635
-0.256066210732154 1.73201355867102
-0.238802466641078 1.72073743888809
-0.221651411560817 1.70947647249609
-0.204612870747546 1.69823547214436
-0.18768661632804 1.68702023941817
-0.170872355277267 1.67583748532089
-0.154169731035583 1.66469467410992
-0.137578301860558 1.65359986122963
-0.121097545216018 1.64256164327732
-0.104726850095554 1.63158911010056
-0.0884655123158842 1.62069179984192
-0.0723127302533912 1.6098796552483
-0.0562676010113092 1.59916298123171
-0.0403291170051192 1.58855240366756
-0.0244961629538079 1.57805882941721
-0.00876751327914871 1.56769340043054
0.00685816974750431 1.557467379504
0.0223823365985671 1.5473920443693
0.0378065504677823 1.53747865787987
0.0531324870122836 1.52773844630513
0.0683619338813362 1.51818257837081
0.0834967900391836 1.50882214503685
0.0985390648893132 1.49966814000397
0.113490877207403 1.49073144094057
0.128354456098494 1.48202279014108
0.143132130711475 1.47355277682603
0.157826341636162 1.46533175358864
0.172439629290407 1.45736977205023
0.186974633045539 1.44967657040419
0.201434088066236 1.44226156830396
0.21582082210104 1.43513386210521
0.230137752227046 1.42830222045724
0.244387881552232 1.42177508023905
0.258574295878865 1.41556054283505
0.272700160331313 1.40966637074567
0.286768718035797 1.40409997967305
0.300783278094319 1.39886839553746
0.314747223982073 1.39397820762063
0.328664000804916 1.38943557289423
0.342537113094313 1.38524622038604
0.356370120554345 1.38141545648125
0.370166633862116 1.37794817030271
0.383930310522356 1.37484883916818
0.397664850776963 1.37212153412247
0.411373993570247 1.36976992554261
0.425061514603098 1.3677972857346
0.438731214106811 1.36620646558343
0.452386924715751 1.3649998711984
0.466032498899096 1.36417947314034
0.479671806697322 1.36374681689979
0.493308731525097 1.36370303360244
0.506947166080776 1.36404885061916
0.520591008361498 1.36478460208069
0.534244157782959 1.36591023929693
0.547910511402911 1.36742534108065
0.561593960243798 1.36932912217986
0.575298385625149 1.37162042412109
0.589027655372882 1.37429770774898
0.602785620063522 1.37735906451398
0.616576109395136 1.38080222941513
0.630402928686721 1.38462459374757
0.64426985550409 1.38882321765566
0.658180636410333 1.39339484249324
0.672138983838896 1.39833590299184
0.686148573087422 1.40364253923818
0.70021303942839 1.4093106074667
0.714335975285711 1.41533568097444
0.72852092740231 1.42171305183729
0.742771394084868 1.42843774287442
0.757090822575198 1.43550452039855
0.771482606547933 1.44290790672761
0.785950083732165 1.45064219245982
0.800496533654629 1.45870144851375
0.815125175502142 1.46707953793542
0.829839166100943 1.47577012747412
0.84464159800986 1.48476669853373
0.859535497705591 1.49406255405891
0.874523823828858 1.50365082558158
0.889609465524037 1.51352448417219
0.904795240890323 1.52367635155462
0.920083895542808 1.53409911097965
0.935478101281142 1.54478531785898
0.950980454863316 1.55572741016205
0.966593476882239 1.5669177185777
0.982319610742773 1.57834847644263
0.998161224106234 1.59001183122094
1.01412059860344 1.60189984730246
1.03019994326744 1.61400452595184
1.04640138447065 1.62631780736609
1.06272696772724 1.63883158112317
1.07917865722957 1.65153769469157
1.09575833546881 1.66442796172386
1.1124678029373 1.67749417013617
1.13777909515582 1.69740536682336
1.146282896308 1.70412148107875
1.16339171162689 1.71766610082143
1.18063669496923 1.73135371461662
1.19801923517476 1.7451761058924
1.21554063903538 1.75912508268328
1.23320213157186 1.77319248372799
1.25100485637061 1.78737018438748
1.26894987597875 1.80165010238484
1.28703817235568 1.81602420336907
1.30527064737946 1.8304845063044
1.32364812340619 1.84502308868717
1.34217134665274 1.85963209427636
1.36084097682153 1.87430373466641
1.37965760445932 1.88903031197005
1.39862174080077 1.90380421428279
1.41773382141698 1.91861792136425
1.43699420714903 1.93346400801409
1.45640318507483 1.94833514731586
1.47596096950753 1.96322411375012
1.49566770302456 1.97812378617852
1.51552345752585 1.99302715070011
1.53552823532103 2.00792730393498
1.55568197027175 2.02281746158028
1.57598452902852 2.03769096628818
1.59643571231175 2.05254128960668
1.61703525620718 2.06736203326063
1.63778283347368 2.08214693034475
1.65867805486257 2.09688984642985
1.67972047044756 2.11158478058317
1.70090957096443 2.12622586630415
1.72224479234975 2.14080737454456
1.74372550436035 2.15532370643193
1.76535103104658 2.16976941383504
1.78712063954111 2.18413919749378
1.80903354474016 2.19842790928053
1.8310889108118 2.2126305515844
1.85328585269782 2.22674227665201
1.8756234376089 2.24075838588525
1.89810068651238 2.25467432909701
1.92071657561258 2.26848570372522
1.94347003782285 2.28218825400624
1.96635996749729 2.29577779577868
1.98938520414178 2.30924938310548
2.01254454061629 2.32259655801807
2.03583669426436 2.33581137573251
2.05926029601146 2.3488845152593
2.08281387713319 2.36180538495609
2.10649585706405 2.37456222509022
2.130304532217 2.38714220744397
2.15423806578444 2.39953153199592
2.17829447849155 2.41171552071146
2.20247164027313 2.42367870847558
2.22676726284574 2.43540493120115
2.25117889314686 2.44687741114585
2.27570390761337 2.45807883947088
2.30033950727207 2.46899145607456
2.32508271361497 2.47959712673411
2.34993036523268 2.48987741758863
2.3748791151795 2.49981366699645
2.3999254290441 2.50938705480014
2.42506558369997 2.51857866903215
2.45029566671035 2.52736957009438
2.47561157636242 2.53574085244471
2.50100902230597 2.5436737038238
2.52648352677232 2.55114946205515
2.55203042634909 2.55814966945169
2.57764487428734 2.56465612486205
2.60332184331739 2.57065093338955
2.62905612895045 2.57611655381723
2.65484235324305 2.58103584377197
2.68067496900203 2.58539210266083
2.70654826440788 2.58916911241293
2.73245636803462 2.59235117605977
2.75839325424488 2.59492315418741
2.78435274893892 2.59687049929346
2.81032853563695 2.59817928808217
2.83631416187421 2.59883625173066
2.86230304588869 2.59882880415961
2.88828848358173 2.59814506834131
2.91426365573201 2.59677390067855
2.94022163544389 2.59470491348722
2.96615539581722 2.59192850172353
};
\addlegendentry{adapt.}
\path [draw=black, line width=1pt, dash pattern=on 9.25pt off 4pt]
(axis cs:0,0)
--(axis cs:0,3.1);

\end{axis}

\end{tikzpicture}

%% file: figures/hw_map_res.tex
\begin{tikzpicture}

\definecolor{darkgray176}{RGB}{176,176,176}
\definecolor{darkgreen}{RGB}{0,100,0}
\definecolor{lightgray204}{RGB}{204,204,204}
\definecolor{maroon}{RGB}{128,0,0}
\definecolor{navy}{RGB}{0,0,128}

\begin{axis}[
width=\textwidth,
height=5cm,
legend cell align={left},
legend style={
  nodes={scale=0.5, transform shape},
  fill opacity=0.8,
  draw opacity=1,
  text opacity=1,
  at={(0.99,0.3)},
  anchor=south east,
  draw=lightgray204
},
tick align=outside,
tick pos=left,
x grid style={darkgray176},
xlabel={X (m)},
xmin=-3.5, xmax=3,
xtick style={color=black},
y grid style={darkgray176},
ylabel={Voxel Size (m)},
ymin=0.2, ymax=0.65,
ytick style={color=black}
]
\addplot [semithick, maroon, mark=*, mark size=1.25, mark options={solid}]
table {%
-3.50186160178299 0.600000023841858
-3.3310454118426 0.600000023841858
-3.13003545887119 0.600000023841858
-2.99883816122154 0.6
-2.80680398293933 0.600000023841858
-2.69764423332597 0.600000023841858
-2.51599857889596 0.600000023841858
-2.3849136763945 0.600000023841858
-2.2314531603864 0.600000023841858
-2.1118675969695 0.600000023841858
-1.98571107815499 0.600000023841858
-1.85561231277486 0.600000023841858
-1.74574337136334 0.600000023841858
-1.65412410321688 0.600000023841858
-1.5608172878006 0.600000023841858
-1.48477997509326 0.600000023841858
-1.40932191877732 0.600000023841858
-1.33732908183076 0.600000023841858
-1.27662696239424 0.600000023841858
-1.23926738362131 0.600000023841858
-1.19661183163215 0.600000023841858
-1.16859621999761 0.600000023841858
};
\addlegendentry{no adapt. 0.6m}
\addplot [semithick, navy, mark=*, mark size=1.25, mark options={solid}]
table {%
-3.52870802195156 0.300000011920929
-3.42585311683462 0.300000011920929
-3.32908345880162 0.300000011920929
-3.19299784809975 0.300000011920929
-3.03878171132756 0.300000011920929
-2.88123042828987 0.300000011920929
-2.72022398284165 0.300000011920929
-2.55572002496491 0.300000011920929
-2.33672335697736 0.300000011920929
-2.21649315860306 0.300000011920929
-2.00689345240581 0.300000011920929
-1.82916155051702 0.300000011920929
-1.68520198531592 0.300000011920929
-1.44839249623364 0.300000011920929
-1.33807442182835 0.300000011920929
-1.17160171508886 0.300000011920929
-0.929453329883819 0.300000011920929
-0.798434137999743 0.300000011920929
-0.573104098607551 0.300000011920929
-0.403688223167938 0.300000011920929
-0.215194109147587 0.300000011920929
-0.0265581667310162 0.300000011920929
0.180991016323148 0.300000011920929
0.369646639612027 0.300000011920929
0.558236895435474 0.300000011920929
0.74674186682676 0.300000011920929
0.935156089458296 0.300000011920929
1.12348420574254 0.300000011920929
1.36820067422987 0.300000011920929
1.4999297611615 0.300000011920929
1.70689009371388 0.300000011920929
1.89500531193344 0.300000011920929
2.08310503880543 0.300000011920929
2.2712000549129 0.300000011920929
2.47810582464853 0.300000011920929
2.62850653135722 0.300000011920929
2.89043956330295 0.300000011920929
};
\addlegendentry{no adapt. 0.3m}
\addplot [semithick, darkgreen, mark=*, mark size=1.25, mark options={solid}]
table {%
-3.44359118521928 0.600000023841858
-3.13530524924011 0.570000052452087
-2.90451208837038 0.550000071525574
-2.80970411664835 0.560000061988831
-2.54243639339841 0.570000052452087
-2.19322172529159 0.550000071525574
-2.06779041650489 0.560000061988831
-1.62198649082142 0.540000081062317
-1.54939577257177 0.550000071525574
-1.22053323500903 0.53000009059906
-1.04130118277996 0.53000009059906
-0.785990290324229 0.510000109672546
-0.586246357956139 0.490000128746033
-0.490802154772139 0.50000011920929
-0.256066210732154 0.510000109672546
-0.137578301860558 0.520000100135803
0.0531324870122836 0.53000009059906
0.186974633045539 0.540000081062317
0.356370120554345 0.550000071525574
0.452386924715751 0.560000061988831
0.616576109395136 0.570000052452087
0.742771394084868 0.580000042915344
0.920083895542808 0.590000033378601
1.04640138447065 0.600000023841858
1.25100485637061 0.600000023841858
1.37965760445932 0.600000023841858
1.72224479234975 0.580000042915344
1.94347003782285 0.560000061988831
2.15423806578444 0.540000081062317
2.3999254290441 0.520000100135803
2.68067496900203 0.50000011920929
2.75839325424488 0.510000109672546
};
\addlegendentry{adapt.}
\path [draw=black, line width=1pt, dash pattern=on 9.25pt off 4pt]
(axis cs:0,0.2)
--(axis cs:0,0.65);

\end{axis}

\end{tikzpicture}

%% file: figures/hw_map_res_cave.tex
\begin{tikzpicture}

\definecolor{darkgray176}{RGB}{176,176,176}
\definecolor{darkgreen}{RGB}{0,100,0}
\definecolor{lightgray204}{RGB}{204,204,204}
\definecolor{maroon}{RGB}{128,0,0}

\begin{axis}[
width=\textwidth,
legend cell align={left},
legend style={
    nodes={scale=0.5, transform shape},
    fill opacity=0.8,
    draw opacity=1,
    text opacity=1,
    at={(0.99,0.01)},
    anchor=south east,
    draw=lightgray204
},
tick align=outside,
tick pos=left,
x grid style={darkgray176},
xlabel={Time (s)},
xmin=25, xmax=43,
xtick style={color=black},
y grid style={darkgray176},
ylabel={Voxel Size (m)},
ymin=0.2, ymax=0.55,
ytick style={color=black}
]
\addplot [thick, maroon, mark=*, mark size=1.25, mark options={solid}]
table {%
24.9903316497803 0.5
25.0925049781799 0.5
25.1810457706451 0.5
25.2788515090942 0.5
25.3929769992828 0.5
25.4847903251648 0.5
25.5786504745483 0.5
25.6792726516724 0.5
25.8067607879639 0.5
25.8816304206848 0.5
26.0041861534119 0.5
26.0780961513519 0.5
26.1817052364349 0.5
26.2800331115723 0.5
26.4033303260803 0.5
26.4771153926849 0.5
26.5770170688629 0.5
26.6933801174164 0.5
26.7816135883331 0.5
26.8877160549164 0.5
26.9843661785126 0.5
27.0849287509918 0.5
27.1765971183777 0.5
27.2813990116119 0.5
27.4100759029388 0.5
27.4772140979767 0.5
27.6104545593262 0.5
27.6768569946289 0.5
27.8071241378784 0.5
27.8783271312714 0.5
27.9905309677124 0.5
28.0784695148468 0.5
28.1766104698181 0.5
28.286096572876 0.5
28.3883354663849 0.5
28.4811151027679 0.5
28.5885784626007 0.5
28.6787574291229 0.5
28.7886729240417 0.5
28.878972530365 0.5
29.000271320343 0.5
29.0826690196991 0.5
29.1776401996613 0.5
29.2782287597656 0.5
29.3846678733826 0.5
29.4953751564026 0.5
29.5832273960114 0.5
29.6941001415253 0.5
29.8011236190796 0.5
29.8774194717407 0.5
29.9903135299683 0.5
30.07772564888 0.5
30.180046081543 0.5
30.2985558509827 0.5
30.3891975879669 0.5
30.4993283748627 0.5
30.5971884727478 0.5
30.6791689395905 0.5
30.7772488594055 0.5
30.8870856761932 0.5
30.9798247814178 0.5
31.0876832008362 0.5
31.1778860092163 0.5
31.2900288105011 0.5
31.3794414997101 0.5
31.4872231483459 0.5
31.5798990726471 0.5
31.6813974380493 0.5
31.7861361503601 0.5
31.8815522193909 0.5
31.9968123435974 0.5
32.0774919986725 0.5
32.1803684234619 0.5
32.2804210186005 0.5
32.3863258361816 0.5
32.4817781448364 0.5
32.59188580513 0.5
32.6767847537994 0.5
32.8113667964935 0.5
32.8767552375793 0.5
32.9922754764557 0.5
33.0853204727173 0.5
33.1787526607513 0.5
33.2879848480225 0.5
33.3797073364258 0.5
33.4833133220673 0.5
33.5861377716064 0.5
33.6920444965363 0.5
33.7786090373993 0.5
33.8852999210358 0.5
33.9850165843964 0.5
34.0772264003754 0.5
34.181672334671 0.5
34.2872440814972 0.5
34.4072864055634 0.5
34.4903161525726 0.5
34.5972480773926 0.5
34.68488240242 0.5
34.7772853374481 0.5
};
\addlegendentry{no adapt. 0.5m}
\addplot [thick, darkgreen, mark=*, mark size=1.25, mark options={solid}]
table {%
27.3557226657867 0.5
27.4037024974823 0.5
27.5095024108887 0.5
27.5970983505249 0.5
27.7160704135895 0.5
27.8027474880219 0.5
27.9128909111023 0.5
28.023508310318 0.5
28.0990793704987 0.5
28.1994333267212 0.5
28.313449382782 0.5
28.3984565734863 0.5
28.5202145576477 0.5
28.5964608192444 0.5
28.792839050293 0.470000028610229
28.9032950401306 0.450000047683716
29.0138132572174 0.430000066757202
29.120151758194 0.410000085830688
29.1927103996277 0.410000085830688
29.2332334518433 0.420000076293945
29.3066964149475 0.430000066757202
29.3971490859985 0.440000057220459
29.5221524238586 0.450000047683716
29.5960342884064 0.460000038146973
29.7053906917572 0.470000028610229
29.7972037792206 0.480000019073486
29.9148824214935 0.490000009536743
29.995335817337 0.5
30.1175701618195 0.5
30.1979751586914 0.5
30.3149282932281 0.5
30.395167350769 0.5
30.5239973068237 0.5
30.5963768959045 0.5
30.7496056556702 0.490000009536743
30.7935011386871 0.5
30.9090938568115 0.5
31.0306737422943 0.490000009536743
31.1619400978088 0.480000019073486
31.2656540870667 0.470000028610229
31.3700437545776 0.460000038146973
31.4734647274017 0.450000047683716
31.5682816505432 0.440000057220459
31.6609020233154 0.430000066757202
31.724650144577 0.430000066757202
31.8517224788666 0.410000085830688
31.9513902664185 0.390000104904175
32.052273273468 0.370000123977661
32.1176710128784 0.370000123977661
32.1949684619904 0.380000114440918
32.3775570392609 0.370000123977661
32.4133207798004 0.380000114440918
32.4961104393005 0.390000104904175
32.5955562591553 0.400000095367432
32.7529883384705 0.380000114440918
32.8472044467926 0.360000133514404
32.9551868438721 0.340000152587891
33.0536022186279 0.330000162124634
33.1570990085602 0.31000018119812
33.2244732379913 0.31000018119812
33.3278756141663 0.31000018119812
33.3923659324646 0.320000171661377
33.5248005390167 0.320000171661377
33.5974078178406 0.330000162124634
33.6973378658295 0.340000152587891
33.7946422100067 0.350000143051147
33.9055700302124 0.360000133514404
33.9942271709442 0.370000123977661
34.1239306926727 0.380000114440918
34.1945748329163 0.390000104904175
34.3111414909363 0.400000095367432
34.4277420043945 0.400000095367432
34.5211880207062 0.400000095367432
34.6219093799591 0.400000095367432
34.694563627243 0.410000085830688
34.795015335083 0.420000076293945
34.9135148525238 0.430000066757202
35.0128817558289 0.440000057220459
35.1565749645233 0.420000076293945
35.1915984153748 0.430000066757202
35.3316655158997 0.430000066757202
35.4547145366669 0.410000085830688
35.5645785331726 0.390000104904175
35.6801674365997 0.370000123977661
35.7980830669403 0.350000143051147
35.9043965339661 0.330000162124634
35.9379532337189 0.340000152587891
35.9924902915955 0.350000143051147
36.1666705608368 0.340000152587891
36.2579553127289 0.330000162124634
36.3568108081818 0.31000018119812
36.4436564445496 0.290000200271606
36.5404577255249 0.28000020980835
36.6118831634521 0.28000020980835
36.6983058452606 0.290000200271606
36.7941355705261 0.300000190734863
36.8918147087097 0.31000018119812
36.9933500289917 0.320000171661377
37.0987963676453 0.330000162124634
37.1878688335419 0.340000152587891
37.3006544113159 0.350000143051147
37.4218633174896 0.360000133514404
37.4868099689484 0.370000123977661
37.5937805175781 0.380000114440918
37.6896798610687 0.390000104904175
37.8076546192169 0.400000095367432
37.891190290451 0.410000085830688
37.9920332431793 0.420000076293945
38.0969574451447 0.430000066757202
38.1829152107239 0.440000057220459
38.2955436706543 0.450000047683716
38.3892147541046 0.460000038146973
38.5396018028259 0.450000047683716
38.6222586631775 0.430000066757202
38.7203125953674 0.420000076293945
38.7992038726807 0.420000076293945
38.9281730651855 0.410000085830688
38.9973437786102 0.410000085830688
39.1310346126556 0.410000085830688
39.1979598999023 0.410000085830688
39.3265833854675 0.410000085830688
39.4054396152496 0.410000085830688
39.5087575912476 0.410000085830688
39.6080033779144 0.410000085830688
39.7426769733429 0.400000095367432
39.7863004207611 0.410000085830688
39.9081106185913 0.410000085830688
40.0012459754944 0.410000085830688
40.106271982193 0.410000085830688
40.2057809829712 0.410000085830688
40.3168067932129 0.410000085830688
40.4264962673187 0.390000104904175
40.5507118701935 0.370000123977661
40.6241321563721 0.350000143051147
40.7243316173553 0.330000162124634
40.8250248432159 0.320000171661377
40.9226794242859 0.320000171661377
41.0020959377289 0.320000171661377
41.1145353317261 0.320000171661377
41.1825432777405 0.330000162124634
41.2906413078308 0.340000152587891
41.3828024864197 0.350000143051147
41.5037035942078 0.360000133514404
41.5826714038849 0.370000123977661
41.7033166885376 0.380000114440918
41.7845237255096 0.390000104904175
41.9000129699707 0.400000095367432
41.9819853305817 0.410000085830688
42.1339001655579 0.400000095367432
42.2031898498535 0.400000095367432
42.2886190414429 0.410000085830688
42.3829197883606 0.420000076293945
42.5138008594513 0.420000076293945
42.5837030410767 0.430000066757202
42.703432559967 0.430000066757202
42.8031876087189 0.430000066757202
42.9069290161133 0.440000057220459
};
\addlegendentry{adapt.}
\end{axis}

\end{tikzpicture}

%% file: content/limitation.tex
The framework assumes that the operator is aiming for the center line a
narrow passage when teleoperating through it. While the forward speed input is
not required, additional gains in performance can be achieved through a pruning
approach to assist with directional inputs~\citep{spitzer_fast_2020}.  Further,
the maximum speed bound calculation applies to holonomic teleoperation
only if a depth sensor is pointed in the sideways direction. When multiple depth
cameras are available on the platform, the maximum velocity in the sideways
direction can be calculated according to \cref{eq:vel_bound_teleop} but for the
$y_\body$ and $-y_\body$ directions.

%% file: content/conclusion.tex
A human operator teleoperating an aerial system in a CSAR scenario may
experience a high cognitive load while modulating the robot's speed to safely
navigate through varying environment clutter.  This paper detailed an approach
for automatic maximum speed modulation for teleoperation of a multirotor in
environments consisting of open, cluttered, and narrow spaces. We couple the
motion primitive design and variable-resolution mapping to create a hierarchical
collision avoidance method that modulates the maximum speed and voxel size of
the local occupancy map simultaneously depending on the environment complexity.
The framework is experimentally evaluated both in simulation and real-world
complex environments, including caves, demonstrating that the speed and map
resolution adaptation yields advantages both in terms of time taken and
ability to complete a task. For future work, we plan to
extend this work to combine with directional input assistance.